# ARSGaussian: 3D Gaussian Splatting with LiDAR for Aerial Remote Sensing Novel View Synthesis

*Yiling Yao[a,b,c], Bing Zhang[a,b,*], Wenjuan Zhang[a] , Lianru Gao[d], Dailiang Peng[a,c], Bocheng Li[a,b], Yaning Wang[a,b], Bowen Wang[a]*

[a] *Aerospace Information Research Institute, Chinese Academy of Sciences, Beijing 100094, China*
[b] *College of Resources and Environment, University of Chinese Academy of Sciences, Beijing 100049, China*
[c] *International Research Center of Big Data for Sustainable Development Goals, Beijing 100094, China*
[d] *Key Laboratory of Computational Optical Imaging Technology, Aerospace Information Research Institute, Chinese Academy of Sciences, Beijing 100094, China*

** Corresponding author.*
*E-mail addresses: yaoyiling22@mails.ucas.ac.cn (Y. Yao), zhangbing@aircas.ac.cn (B. Zhang)*

**Abstract—Novel View Synthesis (NVS) can reconstruct scenes from multi-view images and synthesize novel images from new viewpoints, which provides technical support for tasks such as target recognition and environmental perception. Aerial remote sensing can conveniently capture a wealth of multi-view images with just a few flights. However, the challenges brought by large distances and sparse viewing angles during collection can cause the model to easily produce floaters and overgrowth issues due to geometric estimation errors. This results in low visual quality and a lack of precise geometric estimation capabilities. Therefore, this study presents ARSGaussian, an innovative novel view synthesis (NVS) method for aerial remote sensing. The method incorporates LiDAR point cloud as constraints into the 3D Gaussian Splatting approach, adaptively guiding the Gaussians to grow and split along geometric benchmarks, thereby addressing the overgrowth and floaters issues. Additionally, considering the geometric distortions arising from data acquisition, coordinate transformations with distortion parameters are integrated to replace the simple pinhole camera model parameters to achieve pixel-level alignment between LiDAR point cloud and multi-view optical images, facilitating the accurate fusion of heterogeneous data and achieving the high-precision geo-alignment. Moreover, depth, normal and scale consistency losses are introduced into the regularization process to guide Gaussians toward real depth and plane representations, significantly improving geometric estimation accuracy. To address the current lack of dense airborne hybrid datasets, we have established and released AIR-LONGYAN, an open-source dataset containing a dense LiDAR point cloud (8 pts/m²) and multi-view optical images captured by airborne scanners and cameras in diverse scenes. Experimental results demonstrate that our method achieves more realistic NVS compared to other state-of-the-art (SOTA) methods tailored for large-scale scenes (e.g., VastGaussian, Momentum-GS, CityGaussianV2). On the public UrbanScene3D dataset, our approach attains a PSNR of 26.75 dB, showcasing superior rendering fidelity. Furthermore, it significantly improves geometric structure estimation accuracy, reducing the RMSE from 1.626 m (suboptimal LetsGO baseline) to 0.327 m, marking a 79.88% enhancement in geometric estimation precision. Our self-build AIR-LONGYAN dataset and code will be available at [https://github.com/WenjuanZhang-aircas/ARSGaussian](https://github.com/WenjuanZhang-aircas/ARSGaussian)**

**Keywords— Novel view synthesis, 3D reconstruction, 3D Gaussian Splatting, LiDAR point cloud, Aerial remote sensing.**

## 1. Introduction

Novel view synthesis (NVS) is a challenging task which aims to generate new images from unseen perspectives by learning scene information from captured multi-view images. NVS of large-scale scenes from space-based data is significant for applications such as object recognition, environmental perception, real-scene 3D, and low-altitude economics. In the context of aerial remote sensing, it primarily relies on Oblique photogrammetry [1], [2], which employs multi-view images captured along a designed flight path as input [3]. Through algorithms such as Structure from Motion (SFM) [4], Multi-View Stereo (MVS) [5], and the bundle adjustment principle [6], it can efficiently collect data and reconstruct 3D models of the target area. By projecting 3D information from a specific viewpoint, a novel view image synthesis is achieved [7].

However, this technique relies on texture mapping [8] to depict appearance, neglecting 3D radiometric attributes. As a result, there is a significant difference between the spectral radiance values of the synthesized image and the ground truth [9]. For example, as shown in left red circle in **Fig. 1**, the ground truth (a) appears predominantly yellow, while the newly generated view (b) appears bluish. Meanwhile, it employs triangular meshes [10] to model the scene structure, which can only recover the approximate geometric structure and lacks finer details, this is evident in the right red circle of **Fig. 1**, where the cylinder and its shadow in the ground truth are incorrectly represented as a flat trapezoid.

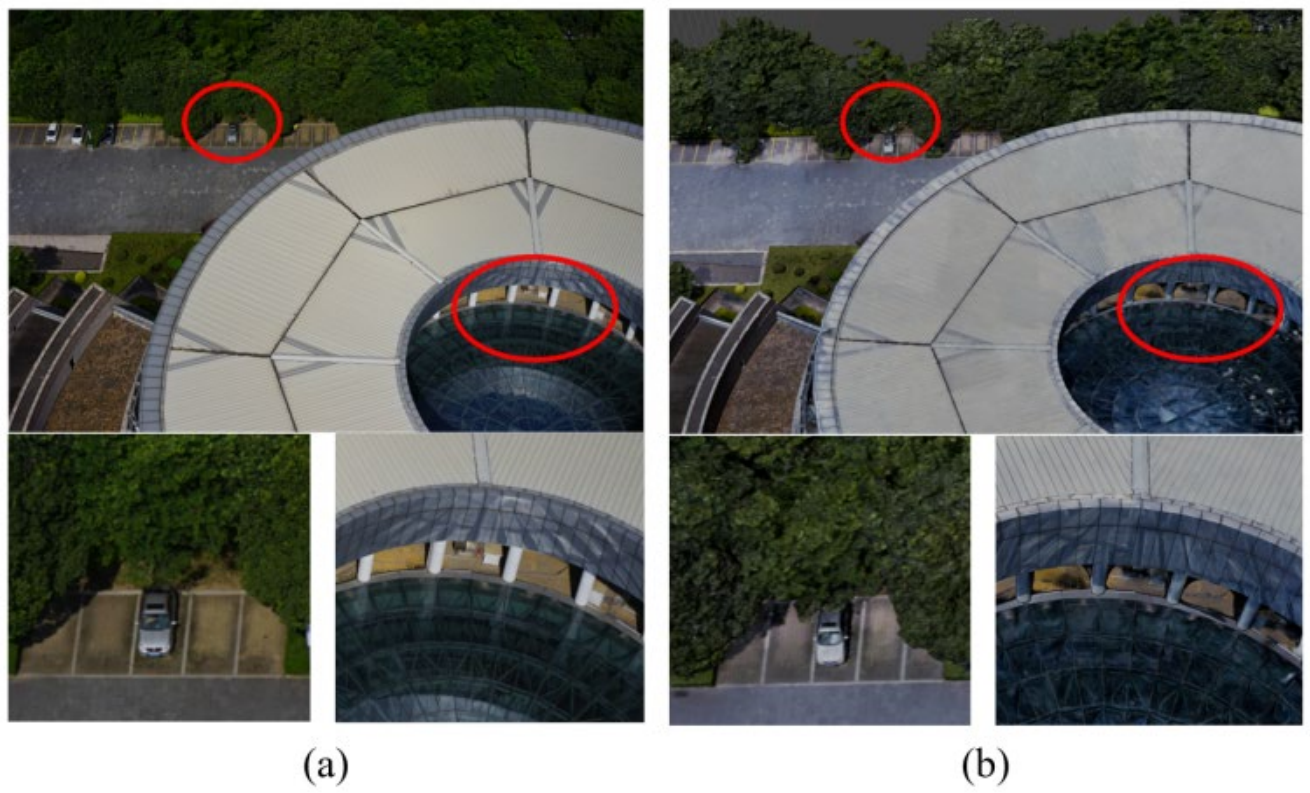

**Fig. 1.** Disadvantages of traditional NVS methods. (a) the Ground truth, (b) Novel view image synthesized by oblique photogrammetry.

In recent years, many innovative novel view synthesis methods have emerged in the fields of computer vision and computer graphics, with the most photo-realistic results coming from Neural Radiance Fields (NeRF) [11] and 3D Gaussian Splatting (3DGS) [12]. NeRF combines neural networks with volume rendering techniques to train an implicit neural radiance field model capable of representing anisotropy of radiation. However, the computational overhead of the volume rendering process is substantial, often resulting in training and rendering times that can span dozens of hours. Despite several optimization efforts [13], [14], [15], [16], NeRF still faces the challenge of slow rendering speed. Additionally, its implicit expression lacks explicit geometric entities and cannot directly express geometric structures. These issues hinder its quantitative application of large scenes [17].

3D Gaussian Splatting (3DGS) [12] has emerged as a significant advancement after NeRF, achieving real-time rendering while producing photo-realistic novel view synthesis. It uses a novel representation called Gaussians to represent scenes, which geometry is represented by Gaussian distributions and appearance is expressed using spherical harmonics to capture the anisotropy of radiation, The main process can be divided into three steps:

1) *Initialization*: 3DGS begins by estimating a sparse point cloud from multi-view images using the SFM algorithm. This serves as the initial geometric framework for the model. Then, Gaussians are grown on the initialized point cloud.
2) *Gaussians Densification*: 3DGS adaptively controls the densification growth of Gaussians to achieve a distribution that performs well in both the general structure and the fine details. For Gaussians whose gradients in the view space exceed a certain threshold, smaller Gaussians are cloned in under-reconstructed regions, or larger Gaussians are split in over-reconstructed regions.
3) *Gaussians Regularization*: During the parameter regularization, the loss function is minimized via backpropagation to optimize the parameters of the Gaussians' covariance matrices (which define the size and shape of the Gaussians) and spherical harmonics (which describe the appearance characteristics of the Gaussians).

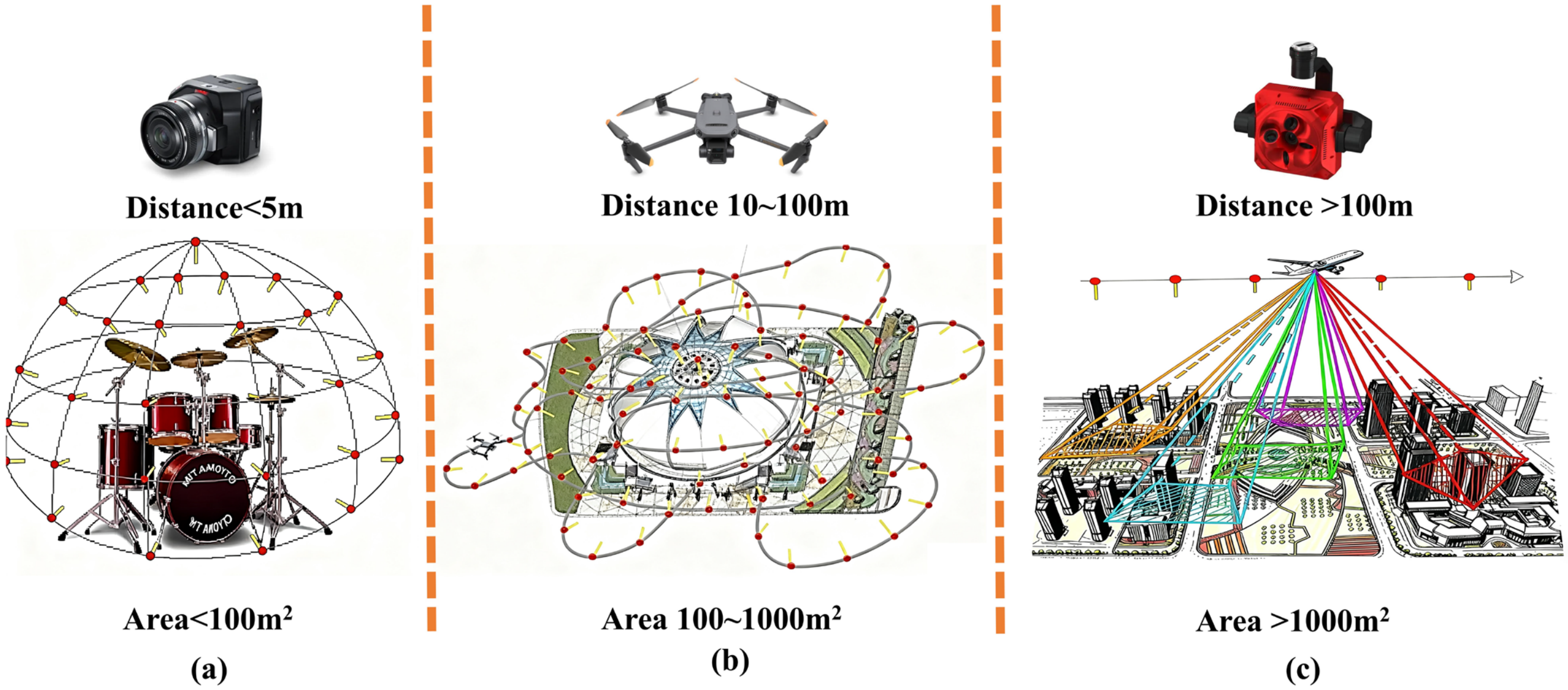

**Fig. 2.** Comparison of different scenes and paths. The collection path is indicated by black lines, and the capture points are marked by red dots. (a) Close-range, small-scale and object-centric scene, (b) large-scale scene, (c) aerial remote sensing scene.

In early research, 3DGS was mainly applied to close-range, small-scale and object-centric scenes( shown in **Fig. 2** (a)), achieving realistic novel view synthesis [18], [19], [20], [21]. However, when the scene extends to a large-scale, such as streets- or city-scale, due to the vastness of the scenery (shown in **Fig. 2** (b)), leading to three primary issues: Firstly, when the scene expands, the data for joint training and the corresponding parameters to be optimized increase dramatically, leading to a decrease in 3DGS efficiency or even insufficient GPU memory[22]. Secondly, as the distance between the camera and the scene increased, the number of effective viewpoints significantly reduced due to the pose limitations of the aircraft, especially in the low zenith angle view. These factors led to errors in depth estimation during the initialization and densification of Gaussians, which caused noticeable floaters and artifacts in the synthesized images [22] as indicated by the red circle in **Fig. 3**. Thirdly, the spatial resolution decreased, and with limited visual prior knowledge and no explicit constraints on true geometric structures, the Gaussians tended to over-stretching in areas which should have geometric consistency during the densification or regularization process [23], as showed by the blue circle in **Fig. 3**. This caused the constructed Gaussians to deviate from the actual structure, leading to bad synthesis results in those areas.

To address these issues, researchers have systematically improved the 3DGS method from aspects such as large-scale scene partitioning and accelerated rendering strategy, Gaussians optimization and geometric constraint enhancement with depth prior fusion. VastGaussian [22] introduces a progressive partitioning strategy to divide the large scene into multiple cells, which ensuring that training images are distributed appropriately based on an airspace-aware visibility criterion, and achieved the application of the first large-scale 3DGS method. Citygaussian [24] improves the partitioning training approach into a divide-and-conquer approach and introduces the Level-of-Detail (LoD) rendering strategy into precision localization and mapping in first-person vehicular dataset.

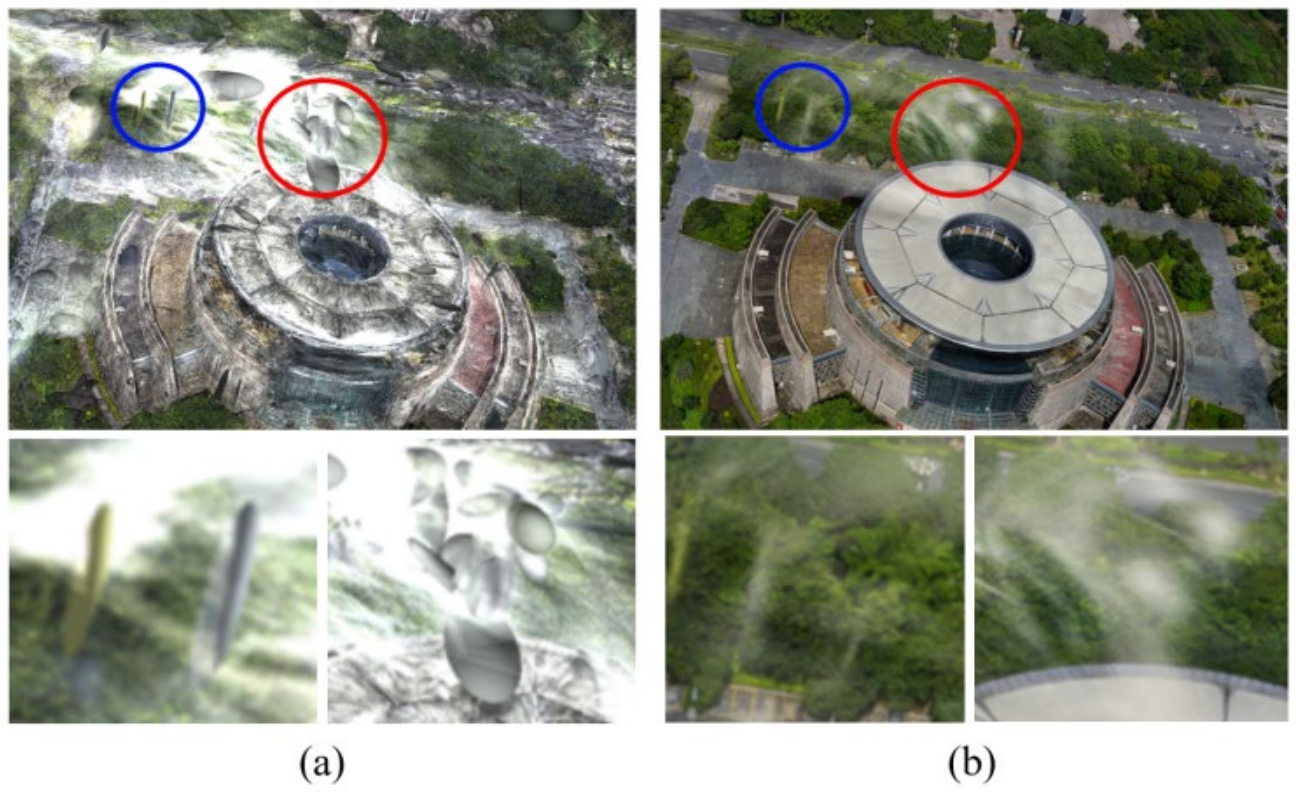

**Fig. 3.** Issues occur in large-scale scenes with 3DGS. (a) Gaussians distribution, (b) Rendered image.

However, aerial remote sensing (shown in **Fig. 2** (c)) differs from typical large natural scenes or first-person Autonomous Driving scenes. In push-broom acquisition paths for aerial remote sensing data, camera poses typically follow highly regular patterns, thus eliminating the need for specialized designs in partitioning training. However, the observation areas are often much broader, and the distance between the sensor and the ground is significantly greater than the effective working range of RGBD cameras [32]. Moreover, the acquisition of aerial optical remote sensing data mainly relies on airborne cameras, which typically provide limited observation angles. For example, the Microsoft UltraCam Osprey [33] and Trimble UX5 HP [34] can only provide five observation angles, with flight heights reaching hundreds of meters; The Airborne Multi-angular TIR/VNIR Imaging System (AMTIS) provides nine observation angles, with flight heights reaching thousands of meters [35]. While the LiDAR

3DGS for the first time, achieving efficient large-scale scene training and rendering. Subsequently, a series of improvements in partitioning strategies and rendering acceleration have been proposed [25], [26], [27]. As for the Gaussians optimization, Pixel-gs [30] implements dynamic scaling of the gradient field based on the distance from the scene to the camera, aiming to suppress the growth of blurring and needle-like artifacts that arise in regions with insufficient initial points. Gaussianpro [23] employs patch-matching techniques to generate normal maps and formulates a progressive Gaussian propagation strategy to guide the densification of Gaussians, thereby improving the suboptimal reconstruction performance of 3DGS in outdoor scenes. As for the geometric constraint enhancement, Gs-slam [31] employs synchronously acquired RGBD cameras as depth priors to control the densification of Gaussians. LetsGo [28] leverages depth information acquired from handheld Light Detection and Ranging (LiDAR) scanners to introduce a depth regularization, effectively eliminating floaters and artifacts, enabling efficient and high-quality NVS in underground garage scene. LiV-GS [29] employs vehicle-mounted LiDAR as a geometric constraint to design a SLAM framework, achieving high-data is also captured by onboard LiDAR scanner. Unlike handheld or vehicle-mounted LiDAR, LiDAR point cloud gaps are common in that case, especially at junctions between the ground and buildings. What's more, aerial remote sensing applications not only require realistic visualization results but also emphasize high-precise quantitative analysis, such as building height measurement, digital elevation model (DEM) construction, and deformation monitoring, all of which require accurate 3D structure and depth information which is lacking in the previous pure visual improvements.

Therefore, the aforementioned large-scale methods are challenging to apply directly to aerial remote sensing applications. LiDAR data is often used as a high-precision geometric prior knowledge to improve the accuracy of reconstruction or positioning in other complex scenes where first-person data is collected [36], [37]. Meanwhile, the integration of LiDAR with multi-view images in aerial remote sensing hardware has been increasing gradually in recent years. For instance, in 2022, Leica introduced the CityMapper [38], a hybrid urban aerial photography system that combines a five-lens oblique camera with LiDAR into a single sensor, enabling the simultaneous acquisition of multi-view images and high-density 3D LiDAR point cloud. However, due to the particularity of LiDAR point cloud data topology, matching based on features and textures is difficult, while the data fusion of the two requires pixel-level alignment, which brings great challenges to data fusion [39].

Based on the analysis above, we propose ARSGaussian, which utilizes the high-precision depth priors provided by LiDAR point cloud. First, we preprocess the acquired LiDAR point cloud through georeferencing, and quality-control filtering. These data are then fused at the pixel level by Colmap-PCD [40] with feature points extracted from optical multi-view images (after sensor distortion correction) using Structure from Motion (SfM). Both modalities are aligned to WGS84 world coordinate system to serve as the model's initialized input. Second, to address the issues of Gaussians floaters and artifacts, we constrain the densification of the Gaussians based on the geometric reference offered by the LiDAR point cloud. Adaptively guiding the growth and splitting of the Gaussian distribution. Finally, to address the issue of over-stretching and further enhance the accuracy of geometric and depth estimation, we incorporate geometric structure losses (including depth, normal, and scale consistency losses) into the regularization process, to encourage the Gaussians to be closer to the true depth and planar maps generated by the LiDAR points. Meanwhile, due to the inefficiency and GPU memory limitations in large-scale remote sensing scene, we employ VastGaussian[22]'s geometry- and visibility-aware tiling strategy to partition the training data.

Moreover, due to the notable scarcity of open airborne datasets utilized in previous 3DGS work that include onboard dense LiDAR data along with multi-view images. We prepared a comprehensive dataset captured by airborne payload ourselves, named the AIR-LONGYAN. LiDAR data in our dataset not only includes building structures but also ground, roads and vegetation that usually appears in remote sensing scenes, with each part achieving a point cloud density of 4-8 points per square meter.

In summary, our work makes the following significant contributions:

1) We designed a 3DGS-based pipeline that is tailored to the characteristics of aerial remote sensing data and can fuse LiDAR and multi-view images as input. This pipeline leverages the high geometric accuracy of LiDAR as a constraint to guide the Gaussians densification strategy and regularization, effectively improving the visual quality and geometric estimation accuracy of novel view synthesize in aerial remote sensing.
2) We developed a pixel-level precision alignment module for LiDAR point clouds and multi-view optical images that establishes coordinate transformation relationships under a distorted camera model and integrates them into the 3DGS baseline's training and feedback pipeline. By employing progressive bundle adjustment, we optimize camera pose estimation to achieve high-precision alignment under 3D space.
3) Considering the lack of open airborne datasets that simultaneously include dense airborne LiDAR and multi-view images, we have released the self-built AIR-LONGYAN dataset. This dataset includes dense airborne LiDAR point cloud along with multi-view images of a remote sensing scene with multiple features.

Experiments have proven that our method effectively suppresses the generation of floaters in aerial remote sensing scenes, achieves photo-realistic novel view synthesis and recovers high-precision geometric structures at the same time. Moreover, the accuracy of geometric estimation in our method surpasses that of other 3DGS-based approaches. Experiments on the open dataset UrbanScene3D and our dataset AIR-LONGYAN demonstrate that our proposed method significantly enhances the performance of 3DGS in aerial remote sensing scenes.

## 2. Related work

### 2.1. *Neural Radiance Field*

The Neural Radiance Field (NeRF) [11] learns the implicit representation of a 3D scene through a 5D neural network, using a multi-view image dataset and the corresponding camera poses as inputs, enabling the reconstruction of 3D scenes or the novel view synthesis via volume rendering without explicit modelling. Owing to its revolutionary neural neural-network-based implementation and realistic synthesis effects, NeRF has become a research hotspot in the fields of 3D reconstruction and novel view synthesis in computer graphics. Lots of efforts have been devoted to enhancing the photo-realistic rendering of NeRF and expanding its applicability to larger scenes. Mip-NeRF [41] proposes cone sampling in place of ray sampling and introduces Integrated Positional Encoding (IPE), significantly enhancing NeRF's detail expression accuracy across varying camera distances and reducing floaters. Building upon this, MipNeRF 360 [42] further extends the application scenario to unbounded 360° scenes. NeRF in the Wild [43] improves the network architecture to mitigate floaters caused by variable illumination and transient occluders.

Subsequent improvements for large-scale and complex scenes began to emerge: Block-NeRF [44] divides large street scene data into blocks, training and optimizing them separately, and then aligning and merging them, enabling NeRF to be applied in street-scale scenes. MegaNeRF [45] further enhances this block-wise approach with foreground-background separation and geometric visibility reasoning for parallel training, achieving better and faster reconstruction results in large-scale drone datasets. BungeeNeRF [46] introduces a progressive NeRF, enabling joint training at different scales, ranging from satellite level to ground level. SatNeRF [47] pioneers the application of NeRF in remote sensing photogrammetry, proposing a ray casting method based on the RPC camera model to generate meter-level digital surface models using multi-view satellite remote sensing images. LiDeNeRF [48] integrates LiDAR data into the NeRF framework to achieve decimeter-level depth estimation in first-person small-scale scenes. DS-NeRF [49] formalizes depth supervision into a loss function, leveraging sparse 3D points from SFM to constrain NeRF training, thereby improving geometry fitting accuracy with fewer input views.

However, despite improvements based on NeRF-related methodologies, the substantial computational and memory overhead associated with volume rendering remains a persistent challenge. Moreover, the continuous implicit representation inherently lacks the expressiveness for precise 3D geometric detail, thereby inherently limiting its ability to capture fine spatial structures.

### *2.2. 3D Gaussian Splatting*

3D Gaussian splatting [12] has introduced a novel scene representation that synergizes Gaussian function and spherical harmonics, leveraging a rasterization-based rendering methodology for the synthesis of scenes derived from multi-view images. Compared to NeRF, this method circumvents the complexities associated with ray sampling strategies and large MLP networks, achieving more realistic novel view synthesis and real-time rendering. Moreover, Gaussians scene expression adeptly represents geometric precision. However, 3DGS faced the same challenge in scaling to larger scenes same as NeRF. The issue of efficiency or insufficient memory of large-scale scene. Simultaneously, 3DGS exhibits a high sensitivity to sampling frequency [18]. When the camera's focal length or the distance between the camera and the object is altered, noticeable floaters and artifacts are observed.

A series of improvements based on partitioning and accelerated rendering strategy have been proposed. VastGaussian [22] implements a progressive scene partitioning strategy that allocates training sets and appearance encoding for large-scale reconstruction based on drone imagery, effectively mitigating floaters occurrences. Citygaussian [24] improves the partitioning training approach into a divide-and-conquer approach and introduces the Level-of-Detail (LoD) rendering strategy into 3DGS for the first time, achieving efficient large-scale scene training and rendering. Hierarchical-GS [26] introduces a hierarchy of 3DGS which offering an efficient LOD solution for efficient rendering of distant content with effective level selection and smooth transitions for very large scenes. Momentum-GS [27] utilizes a momentum-based self-distillation technique to promote the consistency and accuracy among sub-blocks in block training, thereby enhancing the effectiveness of large-scale scene reconstruction. CityGaussianV2 [25] subsequently builds upon CityGaussian by introducing 2D Gaussian Splatting (2DGS) as the primitive for scene representation, alongside improvements in large-scale reconstruction and parallel training methods, specifically targeting enhanced geometric accuracy.

Furthermore, without explicit constraints on true geometric structures, 3DGS tends to excessively proliferate split Gaussians, leading to substantial redundancy in Gaussians and memory consumption. A series of improvements focuses on refining the Gaussians densification strategy or parameter optimization to address this issue. Pixel-gs [30] introduces a dynamic scaling of the gradient field based on the distance from the scene to the camera, aiming to suppress the growth of floaters near the camera. GaussianPro [23] employs patch-matching techniques to generate normal maps and formulates a progressive Gaussian propagation strategy that guides the densification of Gaussians, thereby resolving the suboptimal reconstruction performance of 3DGS in low-texture regions due to its reliance on SFM point cloud initialization—significantly enhancing rendering quality. Compact-3DGS [50] proposes a grid-based neural field for view-dependent color, and compact geometric attribute representation through vector quantization. LightGaussian [51] significantly compresses 3D Gaussian representations by over 15x, compelling the model to eliminate less contributive Gaussians. ABSgs [52] discusses the shortcomings of 3DGS regarding excessive reconstruction of high-frequency details by addressing gradient collision issues during adaptive densification processes while optimizing reconstruction outcomes at high-frequency detail areas.

Meanwhile, another direction of improvement involves leveraging multi-source data as geometric prior guidance or constraints to address this challenge. Gs-slam [31] employs synchronously acquired RGBD cameras as depth priors to control the densification of Gaussians, and facilitate the development of a SLAM system based on 3DGS. LetsGo [28] leverages depth information acquired from handheld LiDAR scanners to introduce a depth regularization, effectively eliminating floaters and artifacts, enabling efficient and high-quality NVS in underground garage scence. LiV-GS [29] employs vehicle-mounted LiDAR as a geometric constraint to design a SLAM framework, achieving high-precision localization and mapping in first-person vehicular datasets. LVI-GS [37] proposed a tightly coupled LiDAR-vision-inertial mapping framework, combined with 3DGS, utilizes the complementary characteristics of LiDAR and image sensors to capture the geometric structure and visual details of 3D scenes.

However, these methods have achieved improvements in natural large-scale scenes or first-person unbounded scenes, but they are not suitable enough for aerial remote sensing scenes. Therefore, we propose the following methods.

## 3. Method

The overview structure of our method is shown in **Fig. 4**. In **Section 3.1**, we introduce a dynamic densification strategy with LiDAR point cloud as geometric constraints, guiding the growth and splitting of Gaussians. In **Section 3.2**, we design a fusion module to achieve pixel-level alignment between LiDAR point cloud and multi-view images, unifying the fused scene under the WGS84 ECEF geodetic coordinate system. This enables the model to acquire real coordinates and scale under world coordinate system while expanding the camera model from a simple pinhole camera model to distorted camera model to reduce biases caused by camera distortion. Finally, in **Section 3.3**, we design a geometric consistency loss in the regularization process, incorporating depth loss, normal loss and scale consistency loss into the loss function. This aids in optimizing the parameters of the 3DGS to better align with real depth and planes, enhancing the accuracy of the model's geometric estimation.

### 3.1. *Adaptive Gaussians Densification Strategy Guided by LiDAR Point Cloud*

To address the generation of numerous floaters and artifacts with inaccurate depth estimation in regions where the input views are sparse, we aim to improve the densification strategy by leveraging the geometric prior provided by LiDAR. LiDAR uses laser diodes to emit short pulsed laser beams, which possess high directionality, monochromaticity, and coherence, allowing them to measure distances with extremely high precision, thereby offering high-precision geometric information and a robust geometric framework [53]. Based on this, we use LiDAR point cloud as initialization and adaptively direct the Gaussians to grow along the direction obtained by projecting their long axis onto the local tangent plane of the LiDAR point cloud [54].

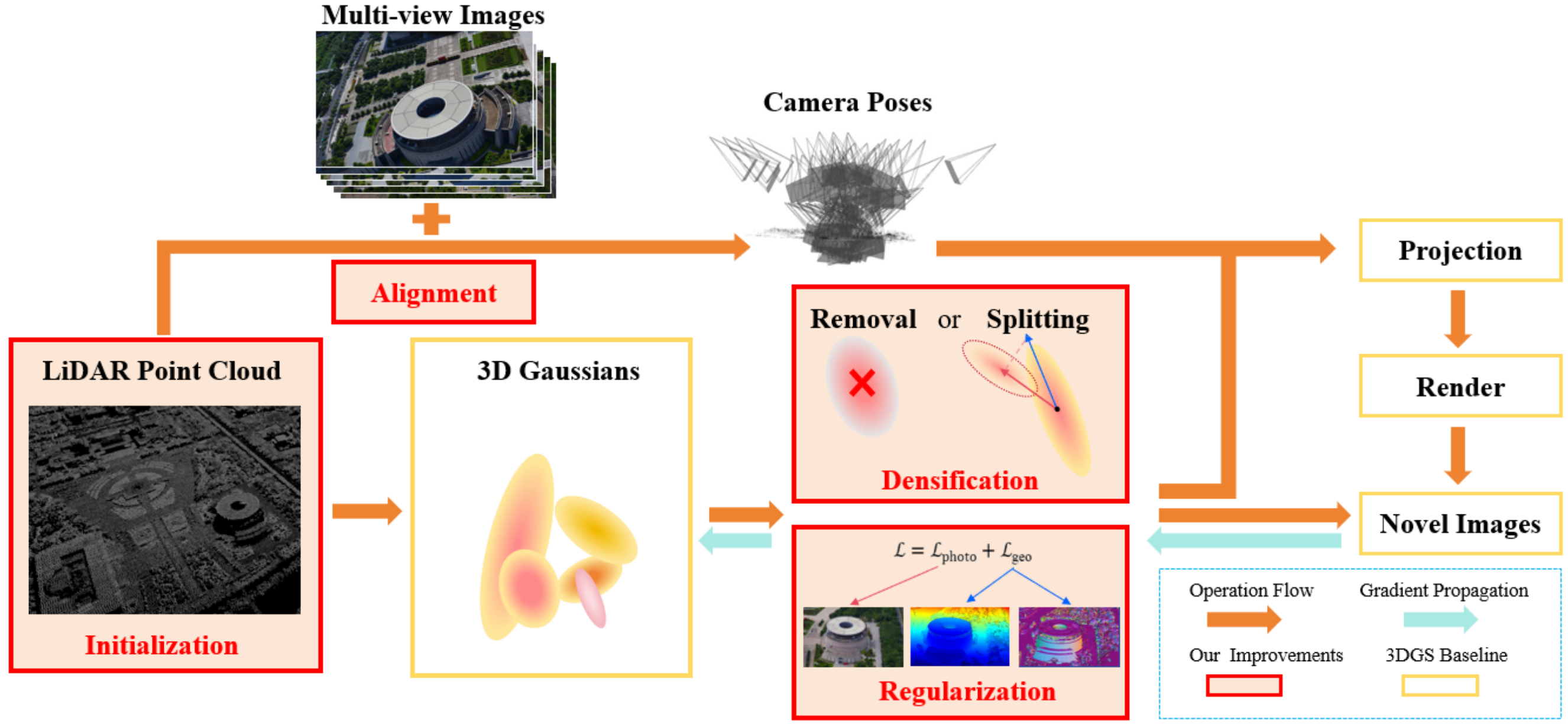

**Fig. 4.** Overview of ARSGaussian

Our densification strategy consists of two parts: determining redundant Gaussians for removal and determining under-reconstructed Gaussians for splitting, and guiding the splitting direction.

### *3.1.1. Removal of Redundant Gaussians*

Define $\mathcal{S} = \{\mathbf{G}_i, \alpha_i, \mathbf{\Sigma}_{3D}^i, \mathbf{c} \mid i = 1 \ldots N_S\}$ as the set of reconstructed 3D Gaussians, where each Gaussian is characterized by its coordinates $\mathbf{G}_i \in \mathbb{R}^3$ in the local level frame coordinate system, its opacity $\alpha_i$, the $3 \times 3$ covariance matrix $\mathbf{\Sigma}_{3D}^i = \mathbf{R}_i \mathbf{S}_i \mathbf{S}_i^T \mathbf{R}_i^T$ that describes the rotation $\mathbf{R}_i$ and scaling $\mathbf{S}_i$ of the Gaussian ellipsoid's three axes, the radiation $\mathbf{c}$ modeled by spherical harmonics, and $N_S$ is the total number of Gaussians in the scene.

By initializing with the original LiDAR point cloud, and at regular intervals defined by a specified number of iterations in the original densification strategy, we employ a Kd-tree [55] nearest neighbor search of Euclidean distance to find the nearest LiDAR point $\mathbf{l}_j = (\mu_{w,lon}^j, \mu_{w,lat}^j, \mu_{w,alt}^j)$, to Gaussians $\mathbf{G}_i = (\mu_{w,lon}^i, \mu_{w,lat}^i, \mu_{w,alt}^i)$ in the local level frame coordinate system, where $(\mu_{w,lon}, \mu_{w,lat}, \mu_{w,alt})$ is the longitude, latitude and altitude of the point in the world coordinate system.

Calculating the plane distance and elevation distance between the two points respectively:

$$d_{plane}^{ij} = \sqrt{\left(\mu_{w,lon}^j - \mu_{w,lon}^i\right)^2 + \left(\mu_{w,lat}^j - \mu_{w,lat}^i\right)^2} \times r_e, \quad (1)$$

$$d_{elev}^{ij} = \sqrt{\left(\mu_{w,alt}^j - \mu_{w,alt}^i\right)^2}, \quad (2)$$

where $r_e$ is the radius of the Earth.

If the distance among two points remains greater than a set distance threshold, indicating that the existing Gaussians have deviated from the correct geometric structure, or if the opacity of the Gaussian point $\alpha_i$ is below a certain threshold $\varepsilon_{opacity}$, signifying that the contribution of the Gaussians is too small, we then delete the current Gaussians.

It should be noted that the remote sensing scene has a certain extent in the planar direction, but a relatively small distance in the elevation direction. Meanwhile, the ranging accuracy of LiDAR is superior to the planar accuracy. Additionally, the LiDAR data obtained by downward-looking from an aircraft is often complete in the horizontal direction but has missing parts in the vertical plane (or areas with horizontal occlusion). In such cases, although there are no LiDAR points near the Gaussians in the vertical area, there are still LiDAR points directly above them, so that the Gaussians still need to be retained. Therefore, it is unreasonable to simply set a single threshold for removal. For these reasons, we separately set a planar threshold $\varepsilon_{plane}$ and an elevation threshold $\varepsilon_{elev}$.

Our strategy can be summarized as follows: Remove Gaussians whose planar and elevation distances from the nearest LiDAR points exceed their corresponding thresholds, and on this basis, still retain the Gaussians that although exceed the threshold but have LiDAR points directly above. This process is illustrated in the part (a) of **Fig. 5**. As shown in **Eq.3**:

$$\begin{gathered} R_i = (P_i \vee E_i \vee O_i) \wedge \neg D_i, \\ P_i = \begin{cases} 1, & if\ d_{plane}^{ij} > \varepsilon_{plane} \\ 0, & otherwise, \end{cases} \\ E_i = \begin{cases} 1, & if\ d_{elev}^{ij} > \varepsilon_{elev} \\ 0, & otherwise, \end{cases} \end{gathered} \quad (3)$$

$$O_i = \begin{cases} 1, & if\ \alpha_i < \varepsilon_{opacity} \\ 0, & otherwise, \end{cases}$$

$$D_i = \begin{cases} 1, & if\ d^{ij}_{plane} < 2cm\ and\ \left(\mu^{j}_{w,alt} - \mu^{i}_{w,alt}\right) > 0 \\ 0, & otherwise, \end{cases}$$

where $R_i$ represents the event that the Gaussian $i$ is removed, $P_i$ represents the event that the Gaussian $i$ exceeds the threshold in the horizontal direction, $E_i$ represents the event that the Gaussian $i$ exceeds the threshold in the vertical direction, $O_i$ represents the event that the opacity of the Gaussian $i$ is below the set threshold, and $D_i$ represents the event that a LiDAR point exists directly above the Gaussian.

### 3.1.2. *Splitting of Under-reconstructed Gaussians*

When determining which Gaussians are under reconstruction, we simultaneously calculate the NDC (Normalized Device Coordinate) coordinates $\left(\mu^{i,k}_{ndc,x}, \mu^{i,k}_{ndc,y}, \mu^{i,k}_{ndc,z}\right)$ of the current Gaussians $i$ under the perspective $k$, we average the gradients of the loss function $\boldsymbol{L}_k$ response used in Pixel-GS [30]. When it is greater than the set threshold $\tau_{pos}$, the existing Gaussians are considered inadequate in fully representing the scene and require densification, represented as:

$$\frac{\sum_{k=1}^{M^i} m^i_k \times \sqrt{\left(\frac{\partial L_k}{\partial \mu^{i,k}_{ndc,x}}\right)^2 + \left(\frac{\partial L_k}{\partial \mu^{i,k}_{ndc,y}}\right)^2}}{\sum_{k=1}^{M^i} m^i_k} > \tau_{pos}, \quad (4)$$

where $M^i$ is all visible perspectives participating in the calculation, $m^i_k$ is the number of Gaussian participates in at viewpoint $k$.

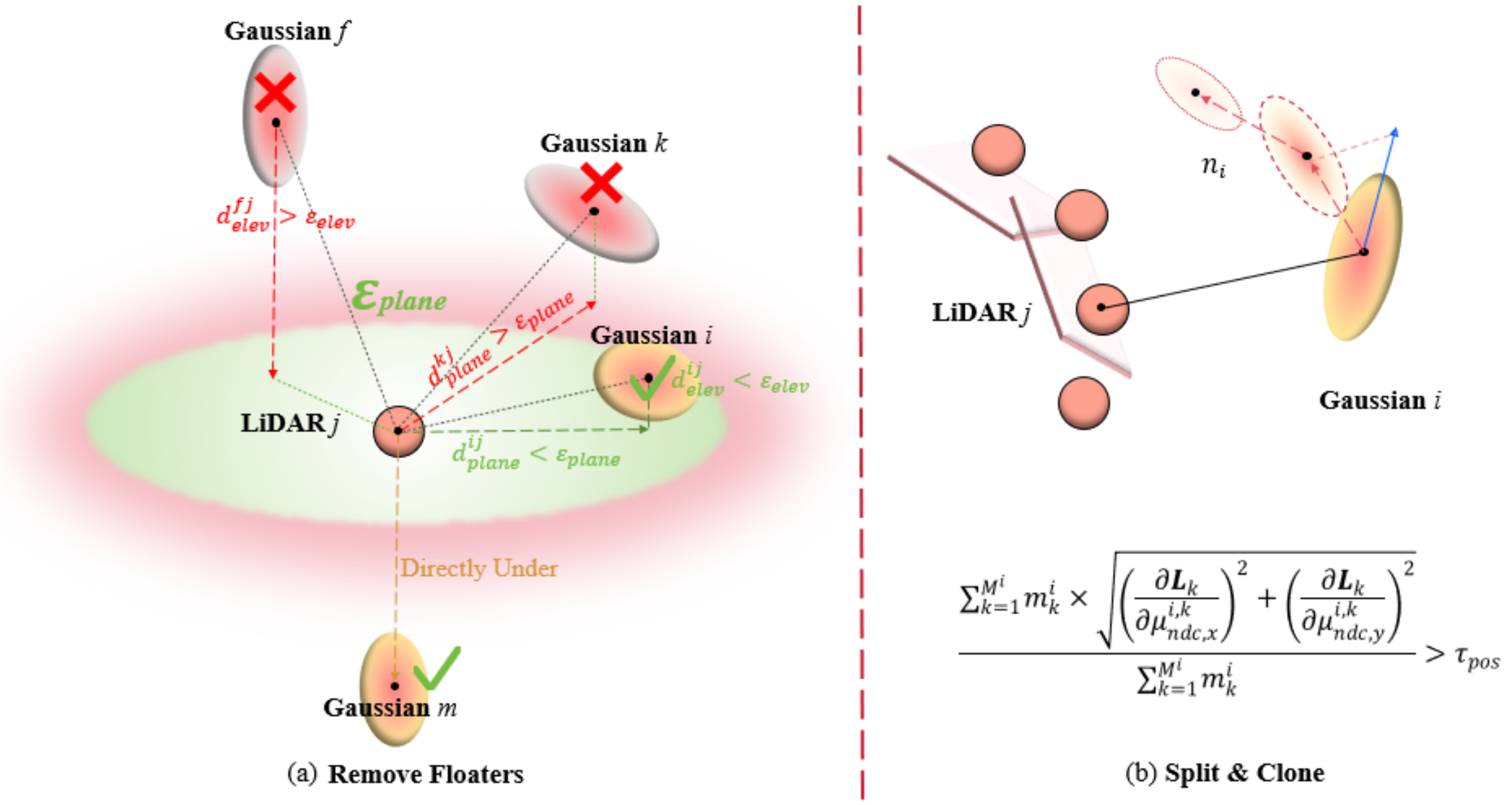

**Fig. 5.** Adaptive Gaussians densification strategy. (a)Removal of redundant Gaussians, (b)Splitting of under-reconstructed Gaussians.

As an ellipsoid-like structure expressed by the covariance matrix, the maximum component of eigenvectors of Gaussians function can be considered as the long axis of the Gaussians and can be written as:

$$n_i = \mathbf{R}_i[r_i, :], r_i = argmax([s_1, s_2, s_3]), \quad (5)$$

where $\mathbf{R}_i$ is the rotation matrix of Gaussians $\mathbf{G}_i$, $[s_1, s_2, s_3]$ is the elements of scaling matrix $\mathbf{S}_i$ of Gaussians $\mathbf{G}_i$, The newly generated Gaussians will grow along the direction obtained by projecting $n_i$ onto the local tangent plane of the LiDAR point cloud, as shown in **Fig. 5** and **Eq.6**:

where N is the local normal plane of the LiDAR point cloud.

Our approach not only fully utilizes the high-precision geometric information from the LiDAR point cloud but also provides a more rational densification direction for Gaussian distributions in 3DGS. Consequently, it effectively alleviates depth errors caused by sparse view angles and insufficient geometric perception.

## 3.2. *Precision Alignment of LiDAR Point Cloud and Multi-View Images*

The application of the method in **section 3.1** requires that the 2D optical multi-view images and the 3D LiDAR point cloud should be precisely aligned. This is particularly challenging for LiDAR data, as it lacks texture features that make direct matching difficult. Direct projection errors can reach dozens of pixels, resulting in poor performance of the synthesis results, especially in areas where height changes dramatically and image distortion edges are evident. This phenomenon is verified in the experiment designed in **Section 4.2.3.2**. To achieve this, we have designed a registration module to achieve pixel-level alignment between LiDAR point cloud and feature points extracted from a 2D image set as followings.

The algorithm flow of this module is illustrated in **Fig. 6**.

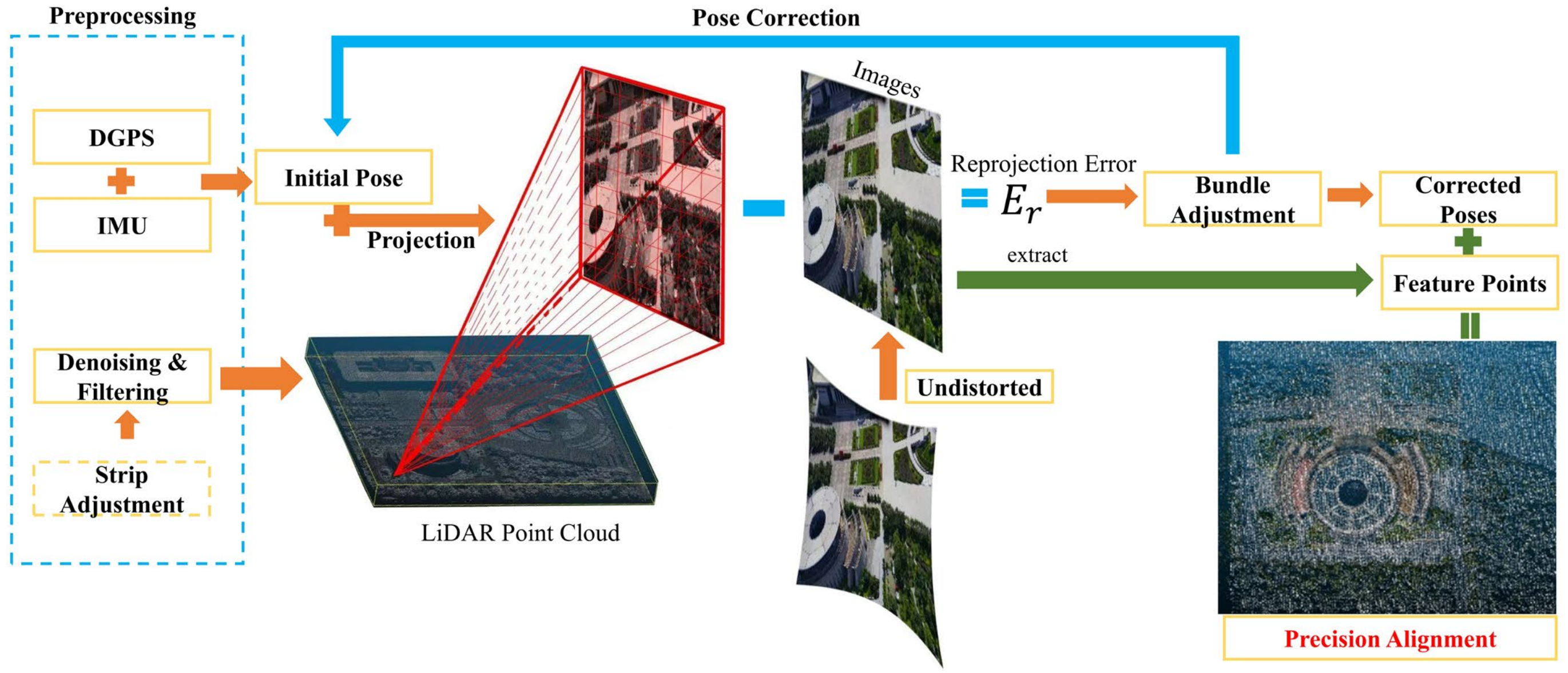

**Fig. 6.** Flow of precision alignment of LiDAR point cloud and images module

### *3.2.1. LiDAR Preprocess*

First, we conducted aerial LiDAR scanning using a laser scanner mounted on a remote sensing aircraft to acquire raw point cloud data of the survey area. Next, we performed differential GPS processing using high-precision GNSS software. By combining airborne GPS observations with ground base station data, we obtained precise WGS84 coordinates of the airborne GPS receiver for each epoch.

Subsequently, we integrated IMU-recorded flight attitude data with the GPS solutions through tight coupling processing to reconstruct the IMU trajectory, including 3D position and orientation at each epoch. Using pre-calibrated lever-arm offsets, we derived the precise trajectory of the LiDAR scanner. Based on the LiDAR trajectory data, we calculated coordinates for each LiDAR point and verified strip alignment accuracy. When necessary, strip adjustment was performed to correct any misalignment.

Finally, we applied statistical outlier removal (SOR) to remove outliers that were either outside the survey area or exhibited abnormal distribution patterns. This included eliminating points with excessive deviations from expected distributions while preserving valid data points meeting project specifications.

For detailed parameter specifications, please refer to section **4.1.1.**

### *3.2.2. World Coordinate Transformation with Distorted Camera Model*

In order to eliminate the image distortion caused by the complexity of the optical system in the remote sensing scenes, so as to ensure the geometric accuracy and the precise alignment, we use the Brown–Conrady camera model [57] which is generally used in the optical system calibration to replace the original simple pinhole camera model, which incorporates the coefficients of the second- and fourth-order terms $k_1$、$k_2$ of radial distortion to accurately describe the optical distortion characteristics of the lens and correct for such distortion. It is capable of calculating radial components by utilizing an even-order polynomial [58], restoring the images to their true geometric relationships.

In pixel coordinate system, define $(u', v')$ as the ideal image coordinates without distortion, $(u, v)$ as the actual image coordinates, $(u_0, v_0)$ as the image coordinates of the optical axis center, and $r$ represent the distance from the image point to the reference point:

$$r = \sqrt{(u' - u_0)^2 + (v' - v_0)^2}. \tag{7}$$

Let $k_1$ and $k_2$ be the second-order and fourth-order distortion coefficients in the respective directions $u$ and $v$. The radial distortion model considering second- and fourth-order lens distortion is given by:

$$\begin{aligned} u' &= u_0 + (u - u_0) \cdot (1 + k_1 \cdot r^2 + k_2 \cdot r^4), \\ v' &= v_0 + (v - v_0) \cdot (1 + k_1 \cdot r^2 + k_2 \cdot r^4). \end{aligned} \tag{8}$$

Therefore, we express the transformation relationship between the pixel coordinate system and the world coordinate system through the distortion camera model as follows:

$$\begin{aligned} \mathrm{Z}\begin{bmatrix} u \\ v \\ 1 \end{bmatrix} = &\begin{bmatrix} 1 + k_1 r^2 + k_2 r^4 & 0 & 0 \\ 0 & 1 + k_1 r^2 + k_2 r^4 & 0 \\ 0 & 0 & 1 \end{bmatrix} \\ &\times \begin{bmatrix} \alpha_x & 0 & u_0 \\ 0 & \alpha_y & v_0 \\ 0 & 0 & 1 \end{bmatrix} \times \begin{bmatrix} f_x & 0 & 0 & 0 \\ 0 & f_y & 0 & 0 \\ 0 & 0 & 1 & 0 \end{bmatrix} \\ &\times \begin{bmatrix} \mathbf{R}_i & \mathbf{t}_i \\ 0^{\mathrm{T}} & 1 \end{bmatrix} \begin{bmatrix} x_w \\ y_w \\ z_w \\ 1 \end{bmatrix} - \begin{bmatrix} u_0(k_1 r^2 + k_2 r^4) \\ v_0(k_1 r^2 + k_2 r^4) \\ 0 \end{bmatrix}, \end{aligned} \tag{9}$$

where $Z$ represents the depth value of the optical center, utilized for the normalization of homogeneous coordinates. $\alpha_x$、$\alpha_y$ denotes the physical dimensions of each pixel in the u-axis and v-axis directions, $f_x$, $f_y$ represents focal lengths in the X and Y directions. After rearranging, we get:

$$\mathrm{Z}\mathbf{l}_{uv} = \mathbf{D}_1 \mathbf{K}(\mathbf{R}_i \mathbf{l}_m + \mathbf{t}_i) - \mathbf{D}_2, \tag{10}$$

where $\mathbf{l}_{uv} = \begin{bmatrix} u \\ v \end{bmatrix}$ denotes the pixel coordinates in the image coordinate system, while $\mathbf{l}_m = \begin{bmatrix} x_w \\ y_w \\ z_w \end{bmatrix}$ stands for the coordinates of points in the LiDAR point cloud within the world coordinate system. $\mathbf{D}_1$、$\mathbf{D}_2$ are distortion matrices, and $\mathbf{K}$ denotes the intrinsic matrix of the camera model.

These matrices are initialized using the combined navigation solution provided by the differential global positioning system (DGPS) and inertial measurement units (IMU). Both the distortion parameters and the intrinsic parameters are provided by the camera calibration file. In **Eq.10**, the feature point coordinates and LiDAR point cloud coordinates are treated as fixed values, serving as inputs for the optimization process to estimate the camera's pose accurately.

### *3.2.3. Estimation of Pose Correction*

To solve the accurate correction of the camera pose, we employ the Colmap-PCD [40] algorithm framework for the solution process. Firstly, the LiDAR point cloud is divided into fixed-size voxels, and a quadrilateral pyramid is formed through four lines. By traversing the LiDAR points within the voxel, for each LiDAR point, we transform its coordinates from the world coordinate system to the camera coordinate system by **Eq.10**. The initial camera poses are provided by the DGPS and IMU, which combines, to solve the initial pose value [56].

Subsequently, we project it onto the imaging plane to obtain its corresponding pixel coordinates in the image $\mathbf{l}_{uv}$, typically represented by multiple pixels. Let $\boldsymbol{\mathcal{F}}^i = \{x_j^i \mid j = 1 \ldots N_{F_i}\}$ be the set of feature points extracted from the image $I_i$, where each feature point is represented by its 2D coordinates $x_j^i \in \mathbb{R}^2$ in the pixel coordinate system and $N_{F_i}$ is the number of feature points. If the feature point $x_j^i$ in the image $I_i$ falls within this range, we regard it as the LiDAR reference point corresponding to the feature point. When multiple LiDAR reference points correspond to a single feature point, we select the one with the smallest distance.

$$\mathbf{e}_r^{ik} = \sqrt{\left(\mathbf{l}_{uv}^k[0] - x_j^i[0]\right)^2 + \left(\mathbf{l}_{uv}^k[1] - x_j^i[1]\right)^2}, \qquad (11)$$

$$\mathbf{E}_r = \sum \mathbf{e}_r^{ik}, k = 1 \ldots N_l,$$

where $l_{uv}^k$ is the pixel coordinates of LiDAR point $k$ after projection, $N_l$ is the total number of LiDAR points.

By minimizing the reprojection error $\mathbf{E}_r$ as shown in **Eq.11**, the optimal estimation of camera poses is solved, thereby achieving pixel-level alignment of LiDAR point cloud and multi-view images in the world coordinate system.

Finally, Colmap-PCD employs three progressive bundle adjustment methods for calculating the optimal estimation of camera poses, incorporating 2D features, image poses, and LiDAR points into a factor graph for optimization [40], achieving precise alignment. The accurately aligned and fused feature points, together with the LiDAR points, serve as the initialization input for model training.

### *3.3. Geometric Consistency Constraint in Regularization*

To address the issue of over-stretching and further enhance the accuracy of geometric estimation, we introduce a geometric consistency constraint into the regularization process of Gaussians parameter optimization called Geoloss. Geoloss utilizes depth maps and normal maps generated from LiDAR point cloud and scale consistency as geometric constraints, which encourage the Gaussians to better conform to the true geometric shape, thereby improving the accuracy and reliability of synthesis results.

The resolution of the LiDAR point cloud from aerial remote sensing is relatively low compared to optical image resolution, which makes it infeasible to directly generate dense depth and normal maps. We therefore first project the LiDAR points set onto the pixel coordinate system according to **Eq.10**, aligning it with the sparse feature points extracted from the RGB images to collectively form the initial planar hypotheses. For pixel depths not covered by LiDAR points, we adopt the assumption from LiDeNeRF [48] and initialize their depth values after Delaunay triangulation interpolation. We then use ACMH [66] which is based on the PatchMatch Stereo [59] for iterative optimization. ACMH further refines the propagation strategy from the GPU-accelerated GIPUMA version [67], enabling more parallel, efficient, and accurate depth map propagation. Through multiple iterations, the depth propagates toward convergence with the ground truth, ultimately yielding a dense depth map and a normal map. These dense maps are subsequently used to compute depth and normal residuals between rendered and real images; these residuals are then adjusted via backpropagation optimization.

The geometric consistency constraint consists of three components: depth constraint, normal constraint, and scale constraint.

$$\boldsymbol{\mathcal{L}}_{geo} = \alpha \boldsymbol{\mathcal{L}}_{depth} + \beta \boldsymbol{\mathcal{L}}_{normal} + \gamma \boldsymbol{\mathcal{L}}_{scale}, \qquad (12)$$

where $\alpha$、$\beta$ and $\gamma$ are all hyperparameters.

The scale regularization loss $\mathcal{L}_{\text{scale}}$ ensures that the shortest axis of the Gaussians can represent the normal direction, thus restricting the minimum scale of the Gaussian distribution to approach zero, effectively suppressing excessive inflation of Gaussians [23].

We represent the depth map and normal map generated from the LiDAR point cloud as the longitudinal direction and planar direction of the scene:

$$\boldsymbol{\mathcal{L}}_{depth} = \sum_{p \in \mathcal{O}} \| \hat{\mathrm{D}}(p) - \bar{\mathrm{D}}(p) \|_1 + \left\| 1 - \hat{\mathrm{D}}(p)^{\top} \bar{\mathrm{D}}(p) \right\|_1, \qquad (13)$$

$$\boldsymbol{\mathcal{L}}_{normal} = \sum_{p \in \mathcal{O}} \| \hat{\mathrm{N}}(p) - \bar{\mathrm{N}}(p) \|_1 + \left\| 1 - \hat{\mathrm{N}}(p)^{\top} \bar{\mathrm{N}}(p) \right\|_1, \qquad (14)$$

where $\hat{\mathrm{D}}$ represents the depth map rendered from the image, $\bar{\mathrm{D}}$ represents the depth map generated by the LiDAR point cloud, $\hat{N}$ represents the normal map rendered from the image, and $\bar{N}$ represents the normal map generated by the LiDAR point cloud. The subscript $p$ denotes the current pixel, and $\mathcal{O}$ denotes the set of all valid pixels.

Ultimately, the geometric constraint can be expressed as a weighted sum of three loss terms, and the overall loss function combines both photometric and geometric losses as follows:

$$\boldsymbol{\mathcal{L}}_{photo} = (1 - \lambda)\mathcal{L}_1 + \lambda \boldsymbol{\mathcal{L}}_{D\text{-}SSIM}, \qquad (15)$$

$$\boldsymbol{\mathcal{L}} = \boldsymbol{\mathcal{L}}_{photo} + \boldsymbol{\mathcal{L}}_{geo}. \qquad (16)$$

Among these, The Structural Similarity Index (SSIM) loss is quantified, denoted as $\boldsymbol{\mathcal{L}}_{D\text{-}SSIM}$ and $\lambda$ is the hyperparameter. This approach not only enhances the accuracy of the reconstruction results but also improves the adaptability and robustness of 3DGS in complex scenes.

## 4. Experiment

### *4.1. Datasets and Implementation Details*

#### *4.1.1. Datasets*

Popular open datasets which simultaneously encompass LiDAR point cloud and multi-view images, such as Waymo [60] and KITTI [61], are predominantly collected using vehicle-mounted LiDAR scanners and cameras, focusing on first-person street scenes and self-driving task. However, these datasets differ significantly from the characteristics of our research scene. As for the aerial-based datasets, among the previously tested datasets in previous 3DGS work, only UrbanScene3D (UR3D) [62] provides synchronized LiDAR point cloud of central building obtained via handheld scanner alongside airborne multi-view images. This dataset has been used in MegaNeRF [45], VastGaussian [22] and Momentum-GS [27]. Consequently, we selected UR3D as the open dataset for testing.

Nonetheless, due to the fact that the LiDAR point cloud in the UR3D dataset is collected using handheld scanner rather than airborne scanner, its validity as a means of verifying aerial data is limited. Additionally, the point cloud density of the UR3D dataset is relatively low, ranging from 0.1-0.3 pts/$m^2$, offering fewer effective geometric constraints in our methodology. Finally, the UR3D dataset only includes LiDAR point cloud of central building, whereas the remote sensing scenes we focus on encompass a more diverse range of features, including ground surfaces, grasslands, vegetation, roads, and other types of objects.

To further demonstrate the effectiveness of our method and its applicability across different scenes, we have also established the AIR-LONGYAN dataset. Our dataset includes high-density point cloud and multi-view images collected using airborne LiDAR scanner and airborne five-cameras, covering a variety of features such as vegetation, plazas, buildings, roads, and grasslands. The average point cloud density in our dataset reaches up to 4-8 pts/$m^2$, which is ten of times higher than UR3D.

It not only fully supports experimental validation for our method but also addresses the scarcity of airborne datasets in related research.

***UrbanScene3D***

UrbanScene3D (UR3D) contains thousands of images under 128k high-resolution covering 16 scenes including large-scale urban regions with 136 $km^2$ areas in total and provides building point cloud acquired by LiDAR scanners.

The LiDAR scanner is Trimble X7 with self-calibration and self-registration techniques. The ranging noise is 0.5mm with a ranging accuracy of 2mm and tan angular accuracy of 21″. The accuracy of 3D points is 1.5mm at 10m and 2.4mm at 20m [62], in the Sci-Art scene we use, the overall error of registration is 3mm. The point cloud density can reach 0.1-0.3 pts/$m^2$.

It is important to note that UrbanScene3D offers datasets collected along various paths. As our focus is primarily on applications of aerial remote sensing, we have selected the dataset collected along the oblique photogrammetric path.

***AIR-LONGYAN***

In the AIR-LONGYAN dataset, we utilized a DJI m300 drone equipped with a high-resolution five-camera array to collect multi-view stereo images of an area of about 325,000 $m^2$ in Longyan City, Fujian Province, following an oblique photogrammetric path of over 400m height. The target area contains buildings, vegetation, roads, and open ground. The dataset comprises 602 images, characterized by a spatial resolution that attains 0.1 meters.

The optical imaging payload is the KT-5, which incorporates 5 full-frame CMOS sensors (1 vertical and 4 tilted at 45 degrees) arranged in a Maltese cross pattern. Each single lens delivers a resolution of $\geqslant$120 million pixels. It employs a downward-mounted configuration with a focal length of 25mm for nadir (straight-down) imaging and 35mm for oblique (angled) imaging. It belongs to multi-lens synchronized frame camera system according to its imaging principle.

The LiDAR payload was the Riegl VQ580-II, mounted on a Cessna 208 aircraft, flying at a relative altitude of approximately 1500 meters. Its repetition frequency is 150 kHz, with a maximum echo recording capability of over ten single pulses. The theoretical point cloud density per flight line was 3.6 pts/$m^2$, with a lateral overlap rate of 69.75%, resulting in an actual point cloud density of 4-8 pts/$m^2$ after three coverage flights and preprocessing. The emitted wavelength was 1550 nm, the pulse divergence angle was 0.25 mrad, the pulse width was 3 ns, the scanning angle range was ±35°. The LiDAR ranging accuracy was 2 mm, the matching accuracy between flight strips is better than 0.2 meters, the positioning error of the airborne POS system is better than 2 cm, so that the overall geometric positioning accuracy was better than 1 meter.

Each photo is accompanied by the initial camera pose provided by the DGPS and IMU, as well as calibrated camera intrinsic parameters. Additionally, control points collected by GNSS RTK equipment were also provided.

#### *4.1.2. Training Details*

Our approach is grounded in the widely recognized open-source baseline of 3DGS-pytorch. Analogous to the training protocol detailed in 3DGS, our model underwent 30,000 iterations across all scenes, adhering to the same training schedule and hyperparameters as 3DGS. Beyond the original cloning and splitting Gaussian densification strategy employed in 3DGS, we integrated our densification strategy introduced in this study, executing it every 300 training iterations. Regarding our loss function parameters, we set $\alpha$= 100, $\beta$= 0.001, $\gamma$ = 0.001, $\lambda$ = 0.2. All experiments were conducted on NVIDIA GTX A6000 GPUs. In all datasets, we use 70% of the data for the training set, 15% for the validation set, and 15% for the testing set.

We evenly divided all scenes into 8 rectangular blocks for parallel training, and extended the edges of each sub-block outward by 30%. According to the camera pose and the visibility after projection (with a threshold set at 70%), we assigned the training set to each sub-block. After the training was completed, we deleted the Gaussians that exceeded the original boundaries of the sub-blocks and performed seamless merging of the entire scene according to Vastgaussian [22].

*4.1.3. Comparison Method & Evaluation Metrics*

We benchmark our approach against state-of-the-art and classical methods in large-scale scene partitioning, accelerated rendering strategies, Gaussians optimization, and geometric constraint enhancement with depth priors. This includes comparisons with the original 3DGS (2023) [12], VastGaussian (2024) [22], Gaussianpro (2024) [23] ,Pixel-GS (2024) [30], LetsGO (2024) [28], Momentum-GS (2024) [27], and CitygaussianV2 (2025) [25].

To evaluate the performance of novel view synthesis, we applied four widely used metrics: Peak Signal-to-Noise Ratio (PSNR), Structural Similarity Index (SSIM), and Learned Perceptual Image Patch Similarity (LPIPS). Additionally, we introduced the Root Mean Square Error (LiDAR RMSE) between LiDAR points and Gaussians across the entire scene as an evaluation metric to measure the geometric accuracy of our method. Finally, we evaluate the efficiency of all methods by comparing their total training time and rendering frames per second (FPS).

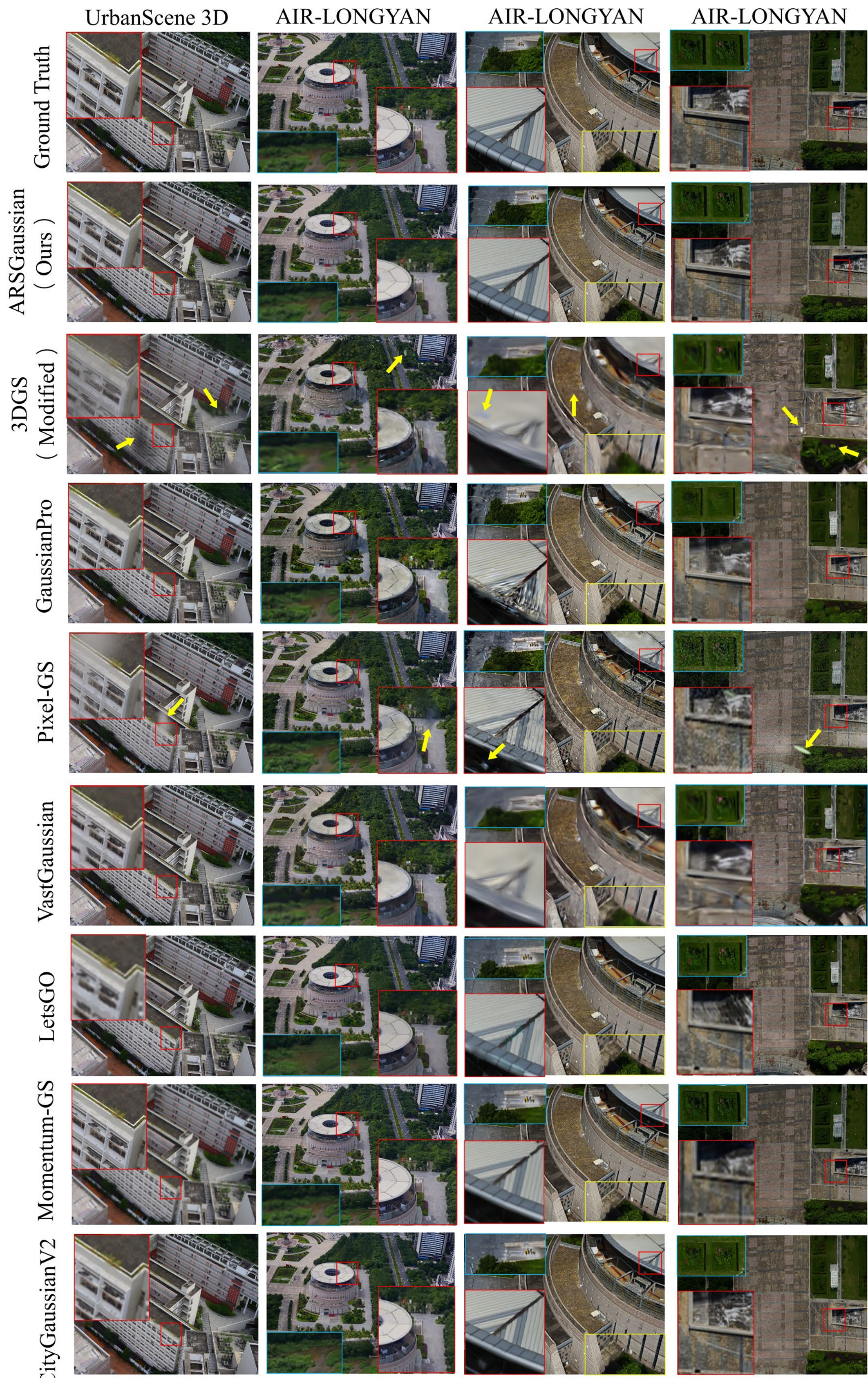

**Fig.7.** Novel view synthesis results between ARSGaussian and previous work. The first line is the Science and Art building in UR3D dataset. The second line is the distance view of the museum building in AIR-LONGYAN dataset. The third line is the close view of the museum building in AIR-LONGYAN dataset.

### *4.2. Results and Discussion*

#### *4.2.1. Visual Result Analysis*

As shown in **Fig. 7**, the floaters are indicated by yellow arrows, the architectural texture details are enlarged by the red box, the vegetative details are enlarged by the blue box, and the near-ground details are enlarged by the yellow box. It can be seen that the original 3DGS produces floaters and artifacts, and is blurred in vegetation and architectural texture details; GaussianPro can effectively inhibit the generation of floaters, but there is still a problem of inconsistent radiation synthesized result in the building texture details which should have obvious geometric consistency; Pixel-GS, due to its targeted improvements, reduces the occurrence of needle-like over-stretching Gaussians, making the vegetative details and architectural textures clearer, but it does not consider the suppression of floaters, there are still a small number of floaters can be observed; VastGaussian, which uses block training to effectively suppress the generation of floaters in large scenes, but at the same time, the total number of training images in each block is reduced due to block training strategy, resulting in lack of clarity and overall blurring in details. Thanks to its targeted improvement for reflective surfaces, LetsGO achieves more realistic NVS results on rooftops with high reflectivity. However, geometric consistency of textures visibly degrades on walls, near-ground areas, and underground garage entrances. This occurs because aerial LiDAR scanning often suffers from point cloud voids or occlusions in these regions, while LetsGO applies uniform processing across all areas. In contrast, both Momentum-GS and CityGaussianV2 maintain consistently clear and photorealistic visual fidelity throughout the scene. However, in vegetated areas, CityGaussianV2 underperforms our method. This occurs because CityGaussianV2 primarily employs 2DGS for rendering to enhance surface reconstruction completeness—an approach that excels for buildings and flat terrain but proves suboptimal for vegetation with complex 3D canopy structures. In contrast, our method strategically constrains only the minor axes of 3DGS via scale factor in our loss function, thereby approximating 2DGS-like compactness while preserving essential 3D characteristics.

To further demonstrate the constraints imposed by our method on floaters, we generated three-view depth distribution maps (set all Gaussian's normalization opacity to 1 and scale to 1, to visualize the relatively transparent floaters) of Gaussians created by 3DGS and our method, as shown in **Fig. 8**. The 3DGS method actually generates a much larger quantity of floaters than expected within the scene, whereas our method effectively constrains these floaters within the true geometric framework.

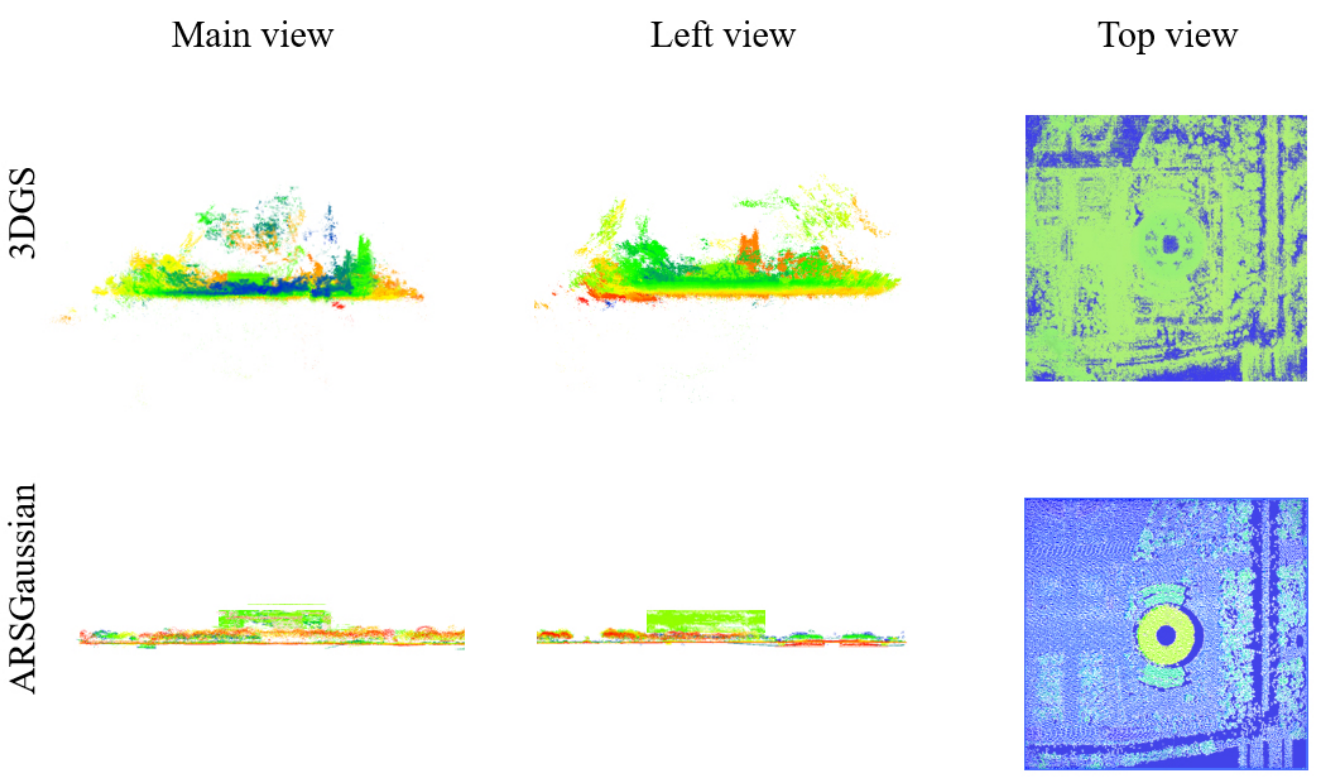

**Fig.8.** ARSGaussian and 3DGS generate three-view distribution of Gaussians positions for the entire scene.

Our method effectively suppressing floaters and achieving clearer synthesized results in both architecture and vegetation area. This improvement is primarily due to the accurate geometry information provided by dense LiDAR point cloud. Additionally, our geometric consistency constraint in regularization process encourages the shape of Gaussians follow the consistency of geometry, avoiding over-stretching. Moreover, our camera model also corrects image distortions, thereby reducing distortions from lens.

However, many state-of-the-art methods have also achieved impressive visual results, necessitating further quantitative analysis.

**Table 1**
Quantitative comparisons of our method compared to previous work on UR3D and AIR-LONGYAN datasets. The best results are in bold and the second results are underlined.

| Dataset | UR3D(Sci-Art) | | | | | |
|---|---|---|---|---|---|---|
| Metrics | PSNR↑ | SSIM↑ | LPIPS↓ | LiDAR RMSE↓ | Training Time↓ | FPS↑ |
| 3DGS | 21.048 | 0.528 | 0.623 | 8.547 | 4h09min | **205.2** |
| VastGaussian | 22.901 | 0.689 | 0.525 | 8.429 | 1h13min | 172.8 |
| Pixel-GS | 24.026 | 0.765 | 0.394 | 7.432 | 4h48min | 92.4 |
| GaussianPro | 25.910 | 0.785 | 0.216 | 5.142 | 3h40min | 42.5 |
| LetsGO (Web) | 23.432 | 0.712 | 0.224 | 1.435 | 1h31min | 56.6 |
| Momentum-GS | 24.115 | 0.857 | 0.304 | 8.514 | 1h12min | 44.7 |
| CitygaussianV2 | **26.874** | 0.864 | **0.215** | 1.823 | **1h04min** | 66.3 |
| **ARSGaussian (Ours)** | 26.747 | **0.876** | 0.218 | **0.783** | 1h28min | 37.2 |

| Dataset | AIR-LONGYAN | | | | | |
|---|---|---|---|---|---|---|
| Metrics | PSNR↑ | SSIM↑ | LPIPS↓ | LiDAR RMSE↓ | Training Time↓ | FPS↑ |
| 3DGS | 21.034 | 0.534 | 0.619 | 8.144 | 3h42min | **205.5** |
| VastGaussian | 22.942 | 0.676 | 0.569 | 8.813 | 1h07min | 171.9 |
| Pixel-GS | 24.639 | 0.783 | 0.337 | 6.837 | 4h22min | 90.5 |
| GaussianPro | 25.722 | 0.823 | 0.204 | 4.216 | 3h16min | 44.6 |
| LetsGO (Web) | 23.106 | 0.708 | 0.231 | 1.626 | 59min | 54.1 |
| Momentum-GS | 25.052 | 0.855 | 0.306 | 8.144 | **43min** | 40.8 |
| CitygaussianV2 | 26.558 | 0.862 | 0.217 | 1.833 | 50min | 66.7 |
| **ARSGaussian (Ours)** | **27.908** | **0.899** | **0.179** | **0.327** | 56min | 36.8 |

#### *4.2.2. Precision Metrics Analysis*

**Table 1** shows quantitative comparison of our method with the comparison methods. It demonstrates our method achieves the best performance in SSIM and LiDAR RMSE on the UR3D open dataset, particularly excelling in depth estimation accuracy. Our method slightly lags behind CityGaussianV2 in LPIPS by only 0.003 and PSNR by only 0.127. On our self-built AIR-LONGYAN dataset, our method attains the best performance in PSNR, SSIM, LPIPS, and LiDAR RMSE, particularly excelling in depth estimation accuracy.

PSNR expresses the difference between images at the pixel level, SSIM expresses the structural similarity between images, as shown in the third row of **Fig. 7**, our method most completely recovers the linear texture structure of the roof. LPIPS uses deep learning method to be closer to human visual judgment, slight numerical differences are acceptable. As such, our method performs outstandingly in quantitatively visual effects, and in terms of geometric estimation precision expressed by LIDAR RMSE, our method has improved 6.97 dB compared with second-best method, greatly improved the accuracy of geometric estimation over all other pure visual methods. All methods utilizing depth priors or geometric enhancements such as LetsGO or CityGaussianV2 achieve some degree of geometric accuracy improvement, but our approach demonstrates the most significant gains. This is because we uniquely employ LiDAR data guided by pixel-level registration in 3D space, whereas other methods merely project generic 3D depth priors to generate 2D depth maps for guidance. Meanwhile, we are also the only method that takes into account the characteristics of aerial remote sensing LiDAR.

To facilitate further analysis, we preliminarily segmented the scene into three distinct categories—vegetation, ground surfaces (including roads), and architecture—for separate quantitative evaluation as shown in **Table 2**. The results indicate that our method exhibits relatively lower accuracy in vegetated areas, primarily due to LiDAR's limited effectiveness in vegetation regions caused by partial canopy penetration and low reflectance. Conversely, geometric accuracy is highest for ground surfaces. However, due to their inherently rough textures, complex patterns, and higher prevalence of transient objects (e.g., vehicles and pedestrians), their visual quality metrics slightly underperform those of architecture regions.

**Table 2** Separate quantitative evaluation of distinct categories.

| Categories | PSNR↑ | SSIM↑ | LPIPS↓ | LiDAR RMSE↓ |
|---|---|---|---|---|
| ground | 27.416 | 0.886 | 0.169 | **0.303** |
| Vegetation | 26.488 | 0.850 | 0.206 | 0.408 |
| Buildings | **27.924** | **0.902** | **0.162** | 0.311 |
| Whole scene | 27.908 | 0.899 | 0.179 | 0.327 |

Additionally, as shown in **Fig. 9**, indicated by the yellow box, the coordinate accuracy of control points reaches the decimeter level (the ground truth provided by GNSS RTK devices). Compared with traditional oblique photogrammetry, our method preserves the precision of control points, while the novel view synthesized results are much closer to the radiation of the ground truth.

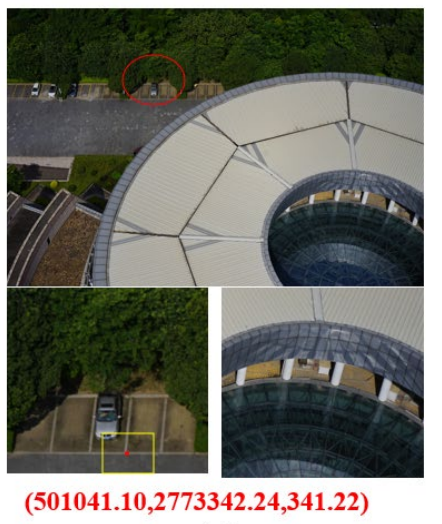

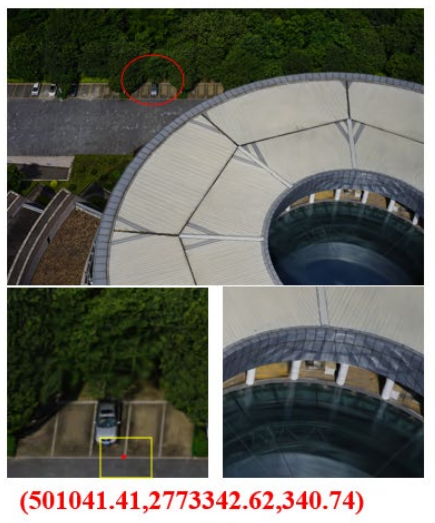

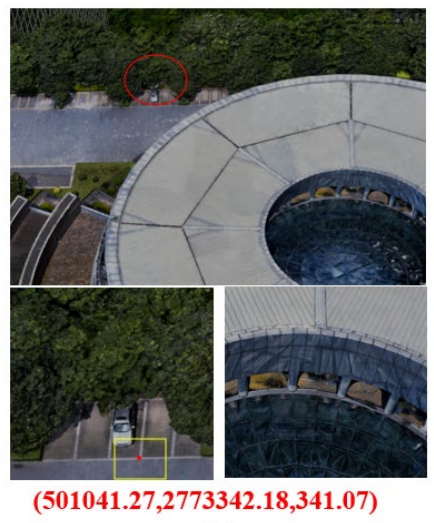

(501041.10,2773342.24,341.22)
(a)

(501041.41,2773342.62,340.74)
(b)

(501041.27,2773342.18,341.07)
(c)

**Fig. 9.** The same perspective comparison, and the coordinates of the control points. (a)the Ground truth, (b)ARSGaussian (Ours), (c)Oblique photogrammetry.

In terms of training efficiency, all block-based methods (VastGaussian, LetsGO, Momentum-GS, CitygaussianV2 and ours) achieve significant reductions in training time. Although our approach also adopts data chunking and block training, it remains slightly slower than purely vision-based methods specifically designed for large-scale scenes with optimized block partitioning and parallel training strategies—primarily due to the computational overhead of precise registration and depth map generation. However, the gap is limited to only tens of minutes. For rendering efficiency, all methods achieve real-time rendering. The original 3DGS pipeline still delivers the highest frame rates, albeit at the cost of sacrificing fine-grained scene details. In contrast, methods incorporating LOD hierarchical rendering (LetsGO and CityGaussianV2) strike a better balance between rendering speed and detail preservation which remains further improvement in our work.

It is noted that the better performance on the self-built dataset may be related to the denser and more comprehensive LiDAR point cloud serving as constraints, a point we will explore further in the ablation study in **Section 4.2.5.1.**

### *4.2.3. Ablation Study*

We performed an ablation analysis on the AIR-LONGYAN dataset to assess the impact of each component in the ARSGaussian model as **Table 3** shows.

**Table 3**
Ablation on LiDAR constraint module, alignment module and geometric loss constraint module.

| Model setting | PSNR↑ | SSIM↑ | LPIPS↓ | LiDAR RMSE↓ |
|---|---|---|---|---|
| w/o LiDARconst | 24.661 | 0.726 | 0.346 | 4.469 |
| w/o Alignment | 20.424 | 0.627 | 0.402 | 1.723 |
| w/o Geoloss | 26.026 | 0.749 | 0.214 | 1.432 |
| Full model | **27.908** | **0.899** | **0.179** | **0.327** |

#### *4.2.3.1. LiDARconst: The Effectiveness of our densification strategy*

To explore the efficacy of our densification strategy applied to 3D Gaussians bound by LiDAR point cloud, we undertook experiments on the AIR-LONGYAN dataset, as illustrated in **Table 3** (first row). Our densification strategy notably augmented various metrics, enhancing both visual quality and the precision of depth estimation. The addition of LiDARconst module mainly improves the LiDAR RMSE metric which means it plays a huge optimization role in the improvement of geometric accuracy.

This enhancement is ascribed to the high-accuracy geometric directives furnished by the LiDAR point cloud. Throughout the densification process, the proliferation and division of Gaussians were meticulously contained within the LiDAR point cloud framework, thereby effectively alleviating problems such as floaters artifacts and overzealous splitting of Gaussians.

#### *4.2.3.2. Alignment: The Necessity of Precise Alignment*

This section aims to elucidate the critical importance of the meticulously devised LiDAR and multi-view images precise alignment scheme discussed previously. As indicated in the second row of **Table 3**, employing solely the LiDAR data post-global orientation as an initialization point for constraining Gaussians densification without the incorporation of precise alignment results in a significant reduction in the PSNR of novel view synthesis, as low as 20.424, even lower than the original 3DGS. Although the geometric accuracy represented by LiDAR RSME is still improved, the visual effects are compromised badly. This phenomenon arises because even minor alignment errors within a few pixels can lead to a complete distortion of depth information, thereby causing disorder in the growth and splitting of Gaussians. This disorder is particularly pronounced in areas with significant height variations, as illustrated in **Fig. 10**, the sculpture at the center of the plaza (marked by a red box) exhibits notable depth changes. Any slight misalignment in these areas would yield severely distorted reconstruction results. Therefore, alignment module is crucial to our method.

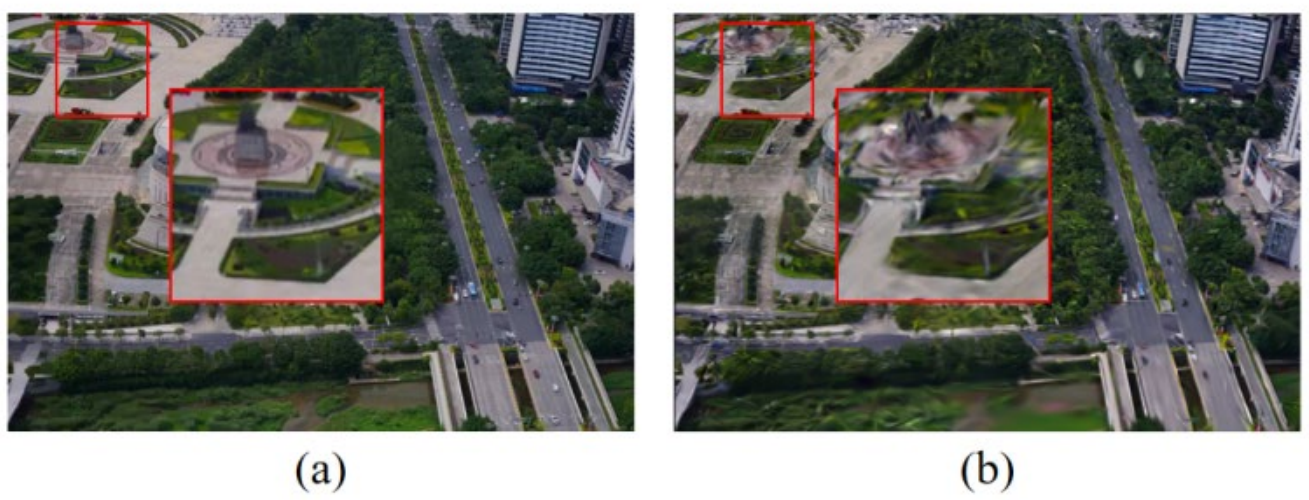

**Fig. 10.** Comparison of precise alignment modules in novel view synthesis. (a) with Alignment, (b) w/o Alignment.

#### *4.2.3.3. Geoloss: Geometric Consistency Constraint*

To investigate the effectiveness of our geometric consistency constraint module, we conducted experiments on the AIR-LONGYAN dataset, as shown in **Table 3** (third row), the Geoloss module further improves the geometric estimation accuracy of the model. That's because the original loss function exclusively considered photometric losses (L1 loss and SSIM loss), prioritizing visual appearance. By incorporating depth, plane and scale consistency losses into the loss function, we achieved enhanced accuracy in geometric estimation while maintaining superior visual quality.

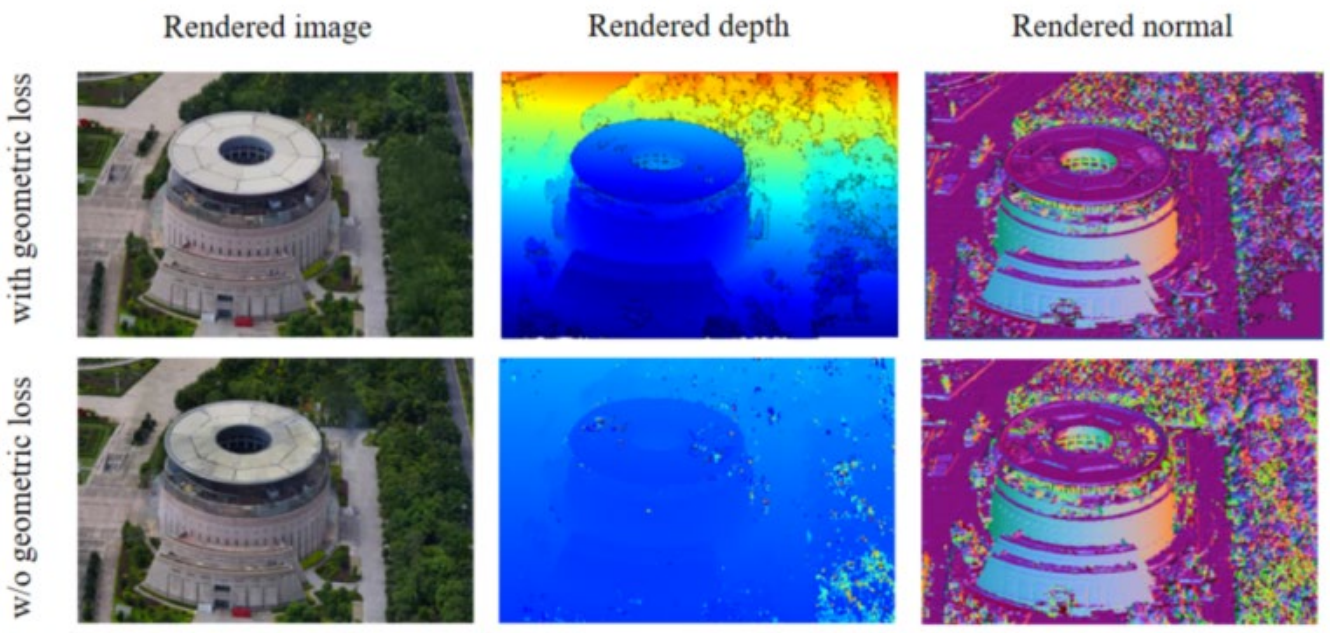

**Fig. 11.** The influence of geometric consistency loss constraints on rendered images, depth maps, and normal maps. (Distance)

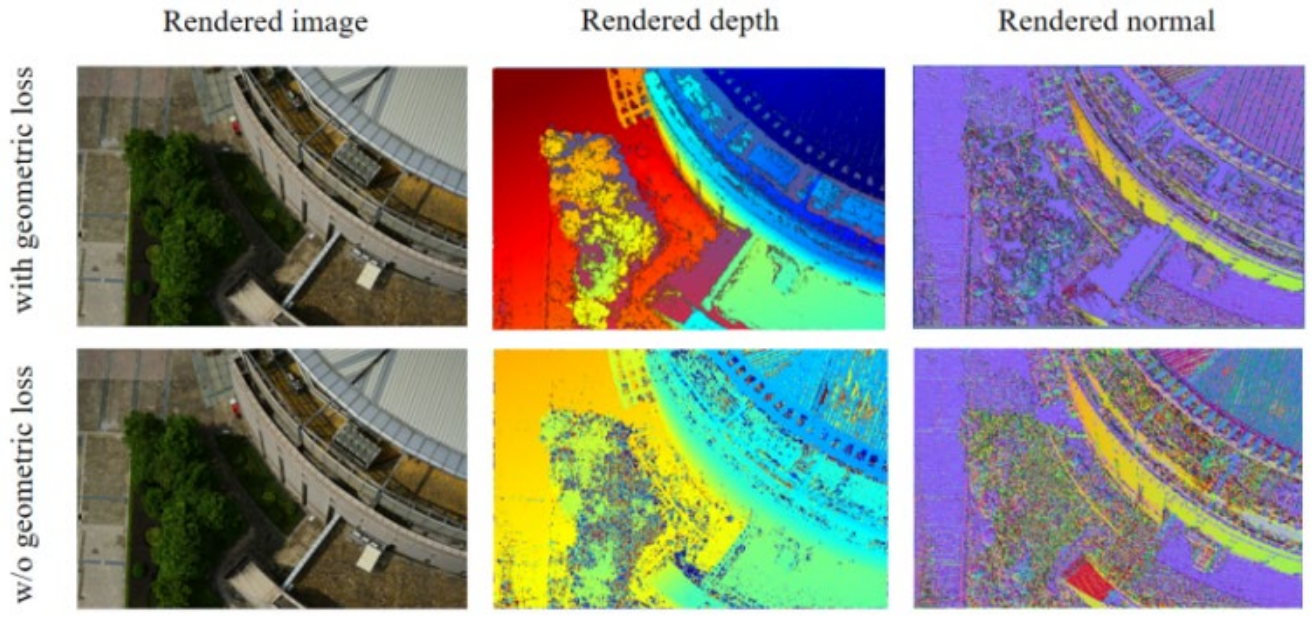

**Fig. 12.** The influence of geometric consistency loss constraints on rendered images, depth maps, and normal maps. (Close)

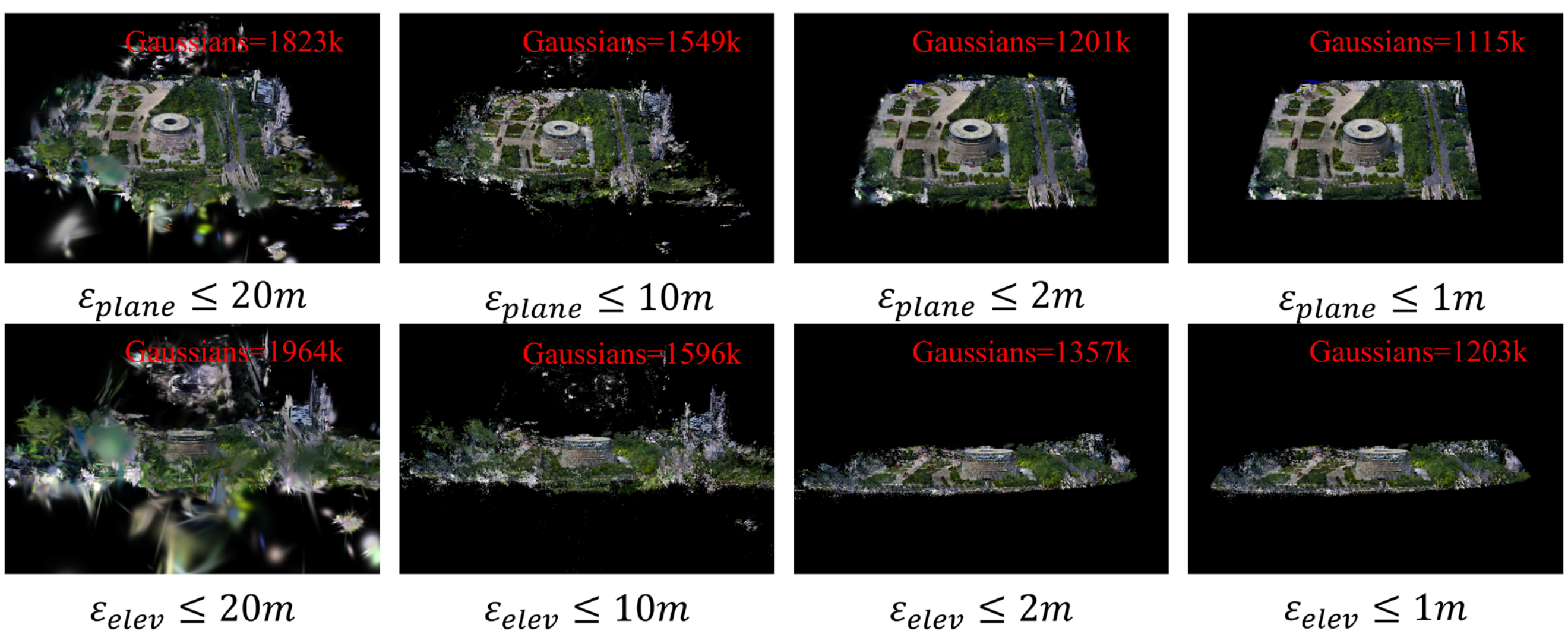

**Fig. 13.** The effect of setting different distance thresholds in densification strategies on the distribution of Gaussians (the number of Gaussians in the scene is marked in red).

**Fig. 11** and **Fig. 12** respectively illustrate the depth maps and normal maps on distance-range and close-range training results. As depicted, incorporating the geometric loss into the loss function only slightly improves the visual quality of the rendered image. However, the introduction of geometric losses notably improved the geometric precision of the rendered depth maps and normal maps.

This improvement is particularly evident in distant view, where rendered depth maps without geometric loss constraints almost fail to express depth differentiation and exhibit substantial noise in vegetated areas. In the close view, it can be observed that the depth and normal direction of the striped structure of the roof maintain geometric consistency after the addition of the Geoloss module.

*4.2.4. Selection of Threshold for Removal*

To explore the optimal value for the threshold set in method **3.1.1**, we conducted a series of comparative experiments with various threshold settings. We set up threshold values of 20m, 15m, 10m, 5m, 2m, and 1m in both plane and elevation directions for the experiments.

As illustrated in **Fig. 13**, the setting of the distance threshold notably impacted the Gaussian distribution. With a progressive decrease in this threshold, the constraints imposed on the splitting and expansion of Gaussians intensified, eventually leading to the convergence of all Gaussians within the LiDAR framework. Although visually there is little difference between setting a threshold of 2m and 1m, from **Table 4** we can see that PSNR values still show a slight improvement.

Although a further reduction in the threshold might theoretically yield numerically superior PSNR values, given that the theoretical geometric positioning accuracy of LiDAR point cloud is 1 meter, a threshold reduction beyond this value would not result in any practical improvement in geometric accuracy. Consequently, we adopted a 1-meter threshold in our final densification strategy.

The total number of Gaussians within the scene are indicated in red. The total number of Gaussians is reduced by pruning irrelevant Gaussians. As a result, fewer parameters need to be optimized. Furthermore, the integration of Gaussian parameter optimization with LiDAR data facilitated the acceleration of the optimization process, enhanced training efficiency, and minimized memory demands (as evidenced in **Table 4**).

**Table 4**
Training Time and Resource Consumption Under Different Thresholds

| Thresholds | PSNR↑ | Gaussians | Training | VRAM |
|---|---|---|---|---|
| 20m | 24.021 | 1823k | 2h03min | 4.49G |
| 15m | 24.574 | 1668k | 1h53min | 3.89G |
| 10m | 25.185 | 1549k | 1h47min | 3.46G |
| 5m | 25.485 | 1378k | 1h32min | 2.25G |
| 2m | 26.247 | 1201k | 1h28min | 1.82G |
| 1m | **27.908** | **1115K** | **56min** | **1.67G** |

*4.2.5. Analysis of the Impact of LiDAR Point Cloud*

*4.2.5.1. LiDAR Density*

The performance discrepancy observed between the UR3D dataset and the AIR-LONGYAN dataset is likely attributable to the higher density of LiDAR point cloud in the AIR-LONGYAN dataset, as illustrated in **Fig. 14**. Notably, the point cloud density in the AIR-LONGYAN dataset significantly exceeds that in the UR3D dataset, which only includes point cloud of building structure while AIR-LONGYAN dataset includes all objects in the region of interest. Consequently, this section will investigate the impact of LiDAR point cloud density on our method. We conducted an experiment by down sampling the LiDAR point cloud in the AIR-LONGYAN dataset to obtain point clouds at densities of 5%, 25%, 50%, and 75%, which served as constraints for the Gaussians densification strategy. The results, presented in **Table 5**, demonstrate that point cloud density significantly influences the quality of novel view synthesis. There is a marked improvement in the quality of synthesized images with increasing point cloud density, peaking at the highest density. Additionally, it is noteworthy that the enhancement in LiDAR RMSE is more pronounced than in the other three metrics as point cloud density increases. This observation aligns with the findings of [63].

This also demonstrates the meaningful value of our open dataset AIR-LONGYAN.

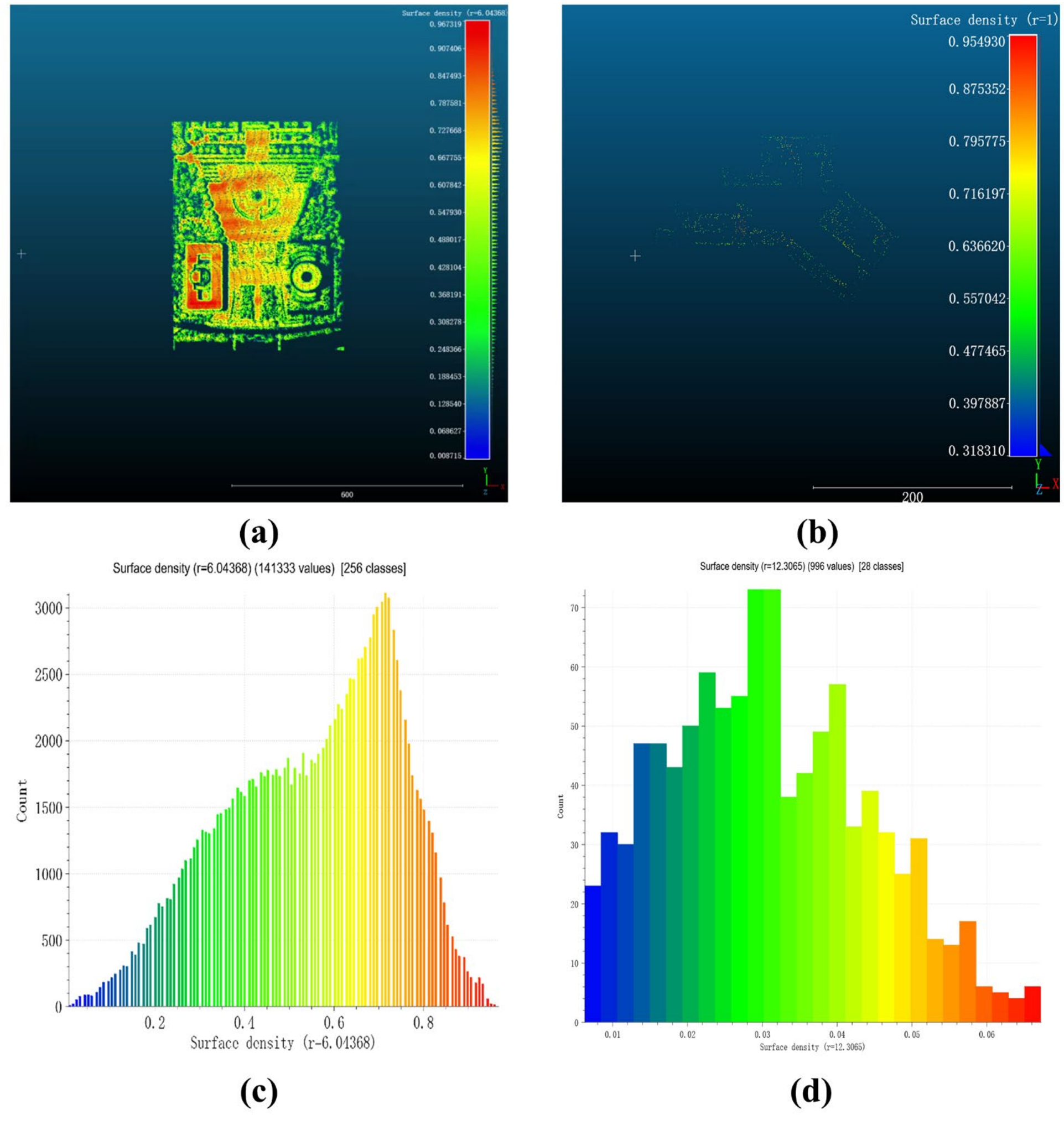

**Fig. 14.** The LiDAR point cloud density maps.

(a) AIR-LONGYAN dataset, (b) UrbanScene3D dataset. (c) AIR-LONGYAN surface density, (b) UrbanScene3D surface density.

**Table 5**
The impact of different point cloud densities on training results.

| Point Cloud Density | PSNR↑ | SSIM↑ | LPIPS↓ | LiDAR RMSE↓ |
|---|---|---|---|---|
| 10% | 26.304 | 0.817 | 0.234 | 0.902 |
| 25% | 26.419 | 0.821 | 0.227 | 0.759 |
| 50% | 26.587 | 0.836 | 0.214 | 0.583 |
| 75% | <u>26.725</u> | <u>0.854</u> | <u>0.202</u> | <u>0.411</u> |
| Dense | **27.908** | **0.899** | **0.179** | **0.327** |

*4.2.5.2. LiDAR Accuracy*

The accuracy of LiDAR point cloud is primarily determined by hardware calibration and ranging precision (millimeter-level), and positioning errors from POS system resolution and atmospheric effects (centimeter-level) [64]. To investigate the impact of LiDAR point cloud accuracy on our method, we superimpose normally distributed random offsets of 2mm, 5mm, 1cm, 2cm, 5cm, and 10cm to each LiDAR point in separate experiments.

As shown in **Table 6**, the LiDAR point cloud accuracy degrades, the visual quality of the model output progressively deteriorates. Beyond 5cm of error, visual metrics experience rapid decline—likely because this threshold exceeds both the LiDAR's calibration tolerance and becomes perceptible to optical imagery. Although thresholds in our densification strategy can tolerant this scale of error, the accumulated point cloud inaccuracies still propagate through misguiding the camera pose in the precision alignment module, and introducing compounding errors in depth and normal estimation.

Notably, the LiDAR RMSE metric shows less pronounced deviation than visual metrics for it is calculated by LiDAR points themselves. Consequently, our method highly relies on LiDAR point cloud precision. Fortunately, standard operational errors—including hardware-level noise (millimeter-scale) and POS positioning errors (<2cm)—are effectively absorbed by both the precision alignment module and densification thresholds. However, severely low-quality point clouds under extreme conditions are not recommended for our method.

**Table 6**
The impact of different point cloud accuracy on training results.

| Point Cloud Accuracy | PSNR↑ | SSIM↑ | LPIPS↓ | LiDAR RMSE↓ |
|---|---|---|---|---|
| **0mm** | **27.908** | **0.899** | **0.179** | **0.327** |
| 2mm | <u>27.902</u> | <u>0.898</u> | <u>0.180</u> | <u>0.329</u> |
| 5mm | 27.517 | 0.892 | 0.183 | 0.335 |
| 1cm | 26.493 | 0.830 | 0.189 | 0.347 |
| 2cm | 24.685 | 0.795 | 0.244 | 0.372 |
| 5cm | 20.118 | 0.547 | 0.457 | 0.452 |
| 10cm | 19.405 | 0.510 | 0.501 | 0.481 |

#### *4.2.5.3. LiDAR Noise Level*

To study the impact of noise levels on our method, we conducted experiments using three types of point clouds: raw unprocessed point clouds, SOR-preprocessed point clouds with different standard deviations σ and neighborhood K values, and preprocessed point clouds with artificially added noise. We simulated three common LiDAR noise categories for artificial noise injection[65]:

*1) Atmospheric noise*

Including backscatter noise and solar radiation noise. When simulating atmospheric noise, the proportion of suspended noise points is set to 5%, with a Z-axis distribution range of 2 - 100 meters and intensity values of 20 – 50 (after normalization); solar radiation noise points account for 2%, with an intensity of 150 - 200, concentrated in the top 30% area; overall, a Z-axis Gaussian noise of 0.2 - 0.5 meters is added (σ = 0.1 - 0.3 meters); in the distant area (> 200 meters), the noise proportion is increased to 10% - 15%, the Gaussian noise σ is increased to 0.08 meters, and Poisson disk sampling (radius 0.5 - 1 meter) is used to ensure a natural distribution of noise points.

*2) System noise*

Including thermal noise and mechanical vibration noise. When simulating system noise, we adopt Gaussian jitter with a standard deviation of 0.01 to 0.03 meters for thermal noise, and set the detector shot noise as a Poisson distribution with an intensity value of 5 to 15 based on the photon counting model; mechanical vibration noise is simulated by sinusoidal disturbances with an amplitude of 0.02 to 0.05 meters (X/Y-axis frequencies of 8 to 15 Hz correspond to common motor speeds, and the Z-axis frequency is halved to reflect the assembly clearance effect), and random white noise of 0.01 to 0.02 meters is superimposed.

*3) Target noise*

Including random noise near targets and transient object motion noise. For the target noise simulation, we set the surface random noise of the processed point cloud in the area with lower intensity as a local Gaussian perturbation of 0.05 to 0.15 meters, and manually add moving targets on the road and in the sky, which are presented as linear streaks of 0.3 to 1.2 meters per second and superimposed with random jitter of 0.1 to 0.4 meters.

The results shown in **Table 7** demonstrate that appropriate denoising effectively enhances the method's NVS performance. However, excessively strict parameter settings (K=100, σ=1.5) may lead to the erroneous removal of valid points, resulting in reduced precision. Among all noise types, the introduction of atmospheric noise has the most significant impact. This occurs because atmospheric noise primarily manifests as suspended particles in airspace, providing nucleation points for floaters and artifacts. Therefore, in the preprocessing, the removal of atmospheric noise needs to be prioritized considered. In contrast, system and target noise are predominantly localized near the point cloud's structural regions, thus exhibiting relatively minor influence on the model.

**Table 7**
The impact of different point cloud noise level on training results.

| Point Cloud noise | PSNR↑ | SSIM↑ | LPIPS↓ | LiDAR RMSE↓ |
|---|---|---|---|---|
| **raw** | 25.237 | 0.612 | 0.311 | 0.626 |
| atmospheric noise | 25.359 | 0.673 | 0.305 | 0.609 |
| system noise | 26.406 | 0.746 | 0.203 | 0.512 |
| target noise | 27.692 | 0.850 | 0.193 | 0.363 |
| After SOR (K=30, σ=2.5) | 27.904 | 0.894 | 0.180 | 0.330 |
| After SOR (K=50, σ=2) | **27.908** | **0.899** | **0.179** | **0.327** |
| After SOR (K=100, σ=1.5) | 27.716 | 0.887 | 0.185 | 0.341 |

#### *4.2.5.4. LiDAR Coverage*

To investigate the impact of point cloud coverage on model performance, we conducted studies by selectively removing portions of the original point cloud and performing dedicated quantitative analysis on the affected viewpoints. Based on common categories of missing LiDAR points, we synthetically generated four types of coverage gaps: glass surface absence (simulating specular reflection failures), asphalt roads absence (modeling low-reflectivity ground surfaces), vertical wall absence (representing occluded facade regions) and Canopy absence (emulating foliage penetration), as shown in **Fig. 15.**

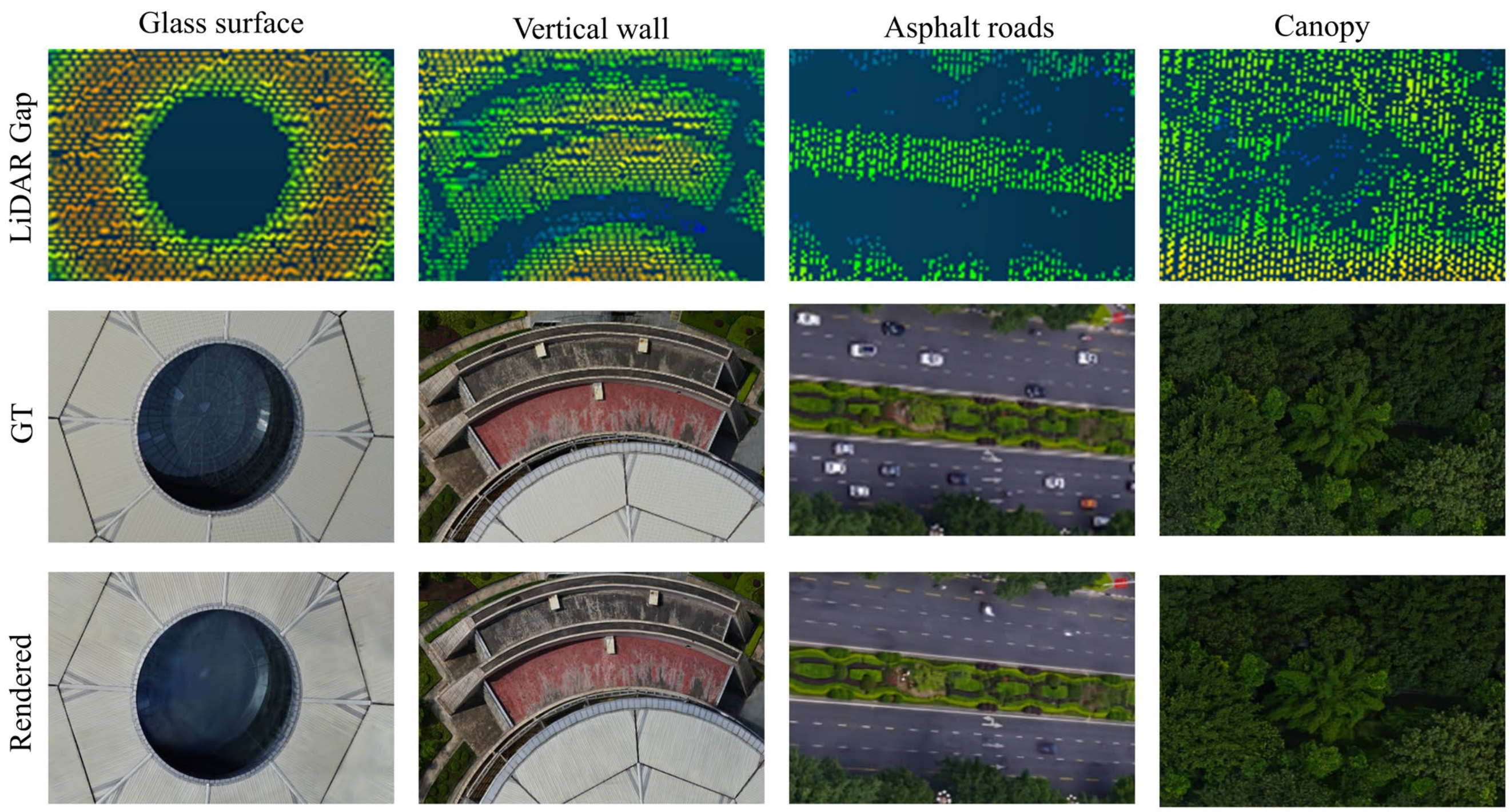

**Fig. 15.** The LiDAR point cloud missing gaps and its corresponding rendered images.

The results show that, as presented in **Table 8**, missing point cloud in glass surface affects both the structure and texture of rendered images, leading to reduced image quality. The absence of point cloud on vertical walls has relatively minor impact on the rendered images. While missing point cloud on asphalt roads also shows significant effects, comprehensive analysis suggests this impact may not only stem from point cloud gaps but also involve excessive transient objects on road surfaces. Canopy gaps result in poorer detail generation for tall trees, with some Gaussians becoming overly stretched. Additionally, due to the complex 3D structure of trees and penetration of LiDAR, these gaps also create the most substantial impact on LiDAR RMSE among all cases.

**Table 8**
The impact of different point cloud coverage on training results.

| Point Cloud Coverage | PSNR↑ | SSIM↑ | LPIPS↓ | LiDAR RMSE↓ |
|---|---|---|---|---|
| **Whole** | **27.237** | **0.898** | **0.180** | **0.329** |
| Glass surface absence | 25.753 | 0.798 | 0.318 | 0.517 |
| Asphalt roads absence | 25.019 | 0.604 | 0.342 | 0.410 |
| Vertical wall absence | <u>27.086</u> | <u>0.882</u> | <u>0.186</u> | <u>0.331</u> |
| Canopy absence | 26.784 | 0.801 | 0.189 | 0.875 |

*4.2.6. Depth Maps Generation Efficiency Analysis*

In our approach, the geometric consistency constraint in regularization relies on a dense depth map from LiDAR. Since generating a dense depth map from LiDAR point clouds is a time-consuming task and our approach is aimed at large-scale scene reconstruction tasks, we propose this section to discuss two critical aspects: efficiency challenges stemming from the time cost of dense depth map generation, and the trade-off between efficiency and performance when comparing different depth map completion algorithms.

We selected several SOTA depth generation methods as baselines to explore the trade-off between efficiency and performance of dense depth map generation in our approach. These methods include sparse depth maps directly projected by LiDAR, the propagation-based depth completion method DOC-DEPTH (2025) [68], the learning-based LiDAR completion method BP-Net (2024) [69], the learning-based monocular RGB image-guided LiDAR depth completion method CHN-Net (2024) [70], the non-learning-based stereo RGB image-guided LiDAR depth completion algorithm SDSGM(2025) [71], the monocular depth prediction method MiDaS(2022) [72] and the multi-view stereo method ACMMP (2022) [66],as shown in **Fig. 16.**

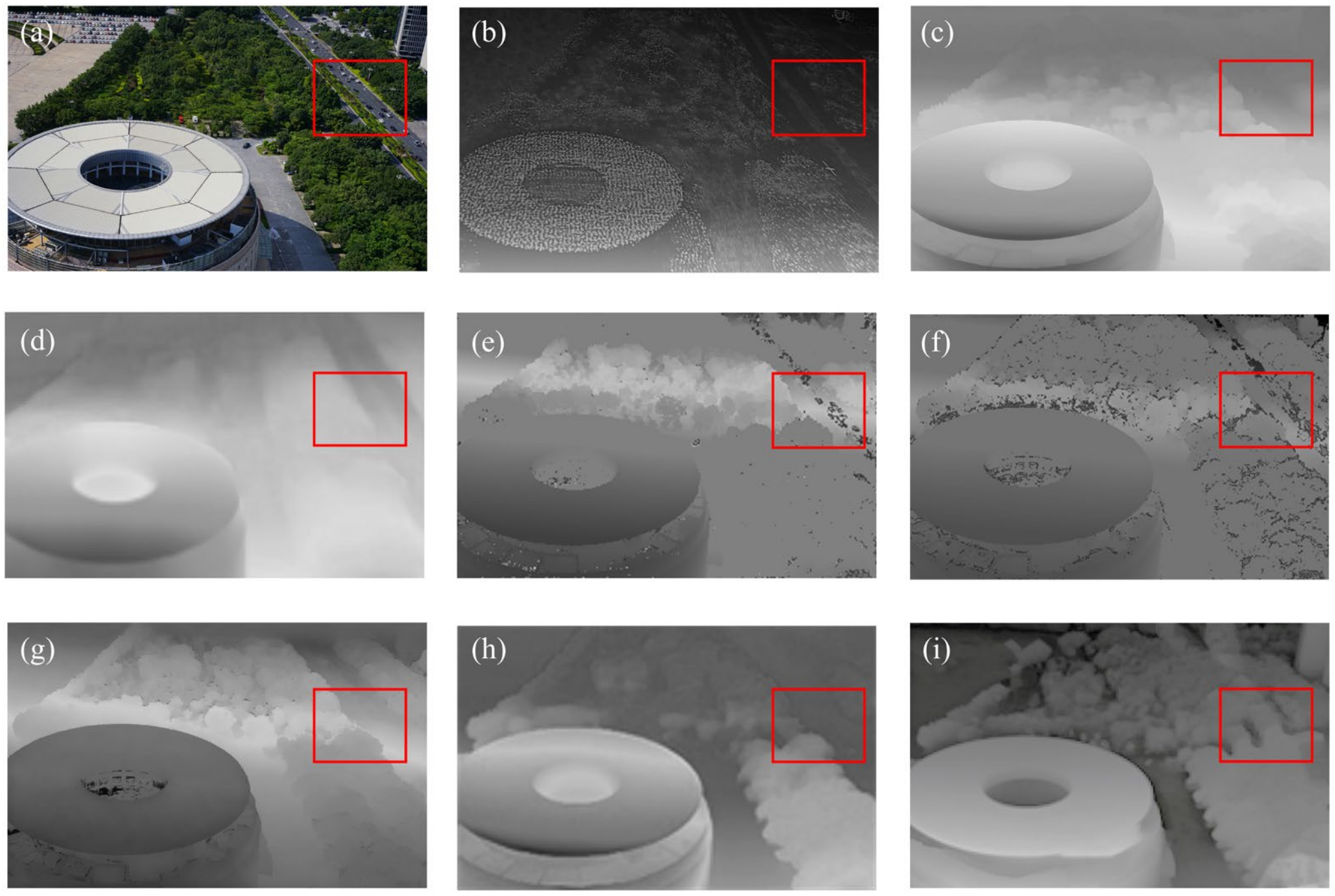

**Fig. 16.** Comparison of different depth map generation methods. (a)RGB image, (b)sparse LiDAR projection, (c) BP-Net, (d) MiDas, (e)Ours. (f)SDSGM, (g)ACMMP, (h) DOC-Depth, (i) CHN-Net

We compared these approaches in terms of average time required to generate a single-frame dense depth map (training time of learning-based methods and SfM time of propagation-based methods are not included), RMSE, and PSNR after used as regularization in our approach, as presented in **Table 9**.

**Table 9**
Comparison of different depth maps generation method results.

| Method | Generation Time (ms) ↓ | RMSE (m) ↓ | PSNR↑ | RGB Guided | Learning Based |
|---|---|---|---|---|---|
| Sparse depth | 0 | 0 | 25.24 | × | × |
| MiDaS | 582 | 8.19 | 26.01 | ✓ | ✓ |
| ACMMP | 2345 | **0.37** | **28.04** | ✓ | × |
| CHN-Net-L | 132 | 1.27 | 26.47 | ✓ | ✓ |
| SDSGM | 68 | 0.86 | 27.46 | ✓ | × |
| DOC-Depth | <u>42</u> | 1.24 | 26.58 | × | × |
| BP-Net | **21** | 2.03 | 26.12 | × | ✓ |
| Ours (10%) | 3293 | 0.78 | 26.30 | ✓ | × |
| Ours (100%) | 786 | <u>0.55</u> | <u>27.91</u> | ✓ | × |

LiDAR-projected sparse depth maps contain numerous missing values (assigned infinite depth) in **Fig. 16**(b). Using them directly for regularization partially corrects the model's inaccuracies but introduces new errors, leading to a drop in PSNR. Monocular depth predictions are blurred, detail-poor, and frequently err in color-varying regions (e.g., the vegetation region) in **Fig. 16**(d) which also lead to a decrease in PSNR. These two types of methods are apparently unsuitable for regularization, so RGB and LiDAR fusion is a more preferable option.

Dense depth maps generated solely from LiDAR input, whether via propagation-based (DOC-Depth, **Fig. 16**(h)) or learning-based methods (CHN-Net, **Fig. 16**(c)), tend to be blurry and lack textural details. In LiDAR-missing regions (e.g., the asphalt road marked with red boxes), they often produce erroneous depth predictions like over-smoothing or assigning very deep values.

ACMMP using geometric and photometric consistency to constrain depth propagation yields dense maps with richer texture, although it has the best result of depth map RMSE and PSNR, its generation time is prohibitively long. High-resolution RGB images exacerbated this computational time, resulting in the total time required to generate depth maps for all images in our dataset reaching up to 4 hours. We recommend that readers who prioritize effectiveness over efficiency adopt this as an alternative.

The learning-based method CHN-Net (guided by monocular RGB images) suffers from significant distortions and noise in **Fig. 16**(i). All learning-based methods are trained on KITTI dataset, a dataset tailored for autonomous driving with specific LiDAR scanners [73]. When transferred to our aerial dataset, their performance degrades sharply. While learning-based completion methods offer fast per-frame rendering, they are severely limited by poor generalization and the days-long training/hyperparameter tuning process.

The non-learning RGB-guided LiDAR method SDSGM retains many gaps, and the dense depth map completeness is insufficient in **Fig. 16**(f). This minor incompleteness is relatively common in non-learning-based depth completion methods. However, overall, it still fulfills its guidance role with high generation speed and depth maps of relatively high quality. Moreover, it avoids the generalization issues inherent in deep learning, making it a viable option for readers who prioritize efficiency.

Our method achieves dense depth completion that balances optical texture details and LiDAR features with relatively low computational cost. Although it has certain flaws in low-texture areas, these flaws do not cause significant losses in the overall model performance. Meanwhile, we also tested when the LiDAR point cloud density was down sampled to 10% as input. The reduction in point cloud density significantly increased both the required time and the number of iterations needed for propagating to generate dense depth maps.

We provide comparisons and analyses against other depth completion methods which allow readers to select a generator based on their trade-off preferences between efficiency and performance.

## 5. Limitation and Discussion

### *5.1. Adaptability of Different Scenes*

Given that urban areas are the primary focus of real-scene 3D applications, our method has been predominantly tested in urban scenes featuring buildings, vegetation, and roads.

However, canopy structure causes multiple scattering of laser light, and only a small fraction of the laser energy can penetrate the canopy to reach the ground. The conventional LiDAR has limited penetration capability for dense canopies, making full penetration difficult.

Meanwhile, due to the selective absorption of laser light at specific wavelengths by water molecules, water bodies attenuate the energy of LiDAR signals. Suspended particles in water cause Mie scattering or Rayleigh scattering, which scatters the laser light in all directions. Owing to laser attenuation and scattering, echo signals become weak with broadened pulse widths, while the point cloud density decreases significantly, further impairing ranging accuracy.

Therefore, these limitations restrict the effectiveness of our method in forest and sea scenes.

### *5.2. Adaptability of Different Aerial Remote Sensing System*

Different aerial remote sensing systems incorporate distinct optical and LiDAR payloads, and the potential limitations inherent in this configuration deserve discussion.

LiDAR systems with varying sensor payloads and acquisition altitudes yield point clouds of differing densities. In this method, airborne LiDAR data with a flight altitude of 1500 m and a repetition frequency of 150 kHz are used. In some lightweight unmanned aerial vehicles (UAV) systems and at lower flight altitudes, the LiDAR point cloud density will increase significantly. Preprocessing through either up sampling or down sampling should be performed based on the actual accuracy requirements, and the spatial resolution of corresponding optical imagery. Overly sparse point clouds may render the current Gaussian densification strategy ineffective, while excessively dense point clouds introduce substantial computational redundancy and memory overhead.

Different remote sensing optical imaging systems also have inherent limitations. Our method uses the Brown–Conrady model to describe distortion for pixel-level alignment, and this camera model performs well in the datasets tested in our method. Since these datasets were collected using oblique multi-lens systems, the Brown–Conrady model effectively captures the distortion differences between such optical sensors and LiDAR payloads. Additionally, because the data was acquired over a short period, variations in environmental illumination and atmospheric conditions were not considered.

However, common airborne remote sensing systems also employ scanning imaging (e.g., whiskbroom imaging), which requires image motion compensation (IMC) [74]. In frame-based imaging systems, there are also airborne composite large-format wide-swath area-array cameras that demand additional block-wise polynomial rational polynomial coefficients (RPC) to model distortion [75]. For acquiring common airborne multi-angle remote sensing datasets: in addition to the simultaneous frame-based imaging with oblique multi-lenses used in our method, there are also systems that use a single lens mounted on the UAV carrier to conduct circumferential observations of the target and capture multi-angle images at different times. For these cameras, extra consideration must be given to motion distortion and non-uniform sampling compensation. What's more, some common aerial cameras, such as the DMC II or DMC III, also require additional parameters to more comprehensively characterize optical distortion [76][77].

When flying at high altitude with a long atmospheric path and unstable conditions, atmospheric distortion should also be accounted for.

### *5.3. Limitation of Efficiency*

Efficiency is also critical for large-scale applications, especially since remote sensing images typically exceed natural scene images in both width and spatial resolution.

Current work, which uses visibility rectangular block partitioning to parallel accelerate training, lacks dedicated efficiency optimizations for training efficiency. Regarding high-speed rendering strategies for large-scale scenes, our experimental analysis found that adopting LoD-based layered rendering methods (e.g., LetsGo and CityGaussianV2) achieves a better balance between rendering speed and detail preservation—there remains room for further enhancement in this aspect within our research.

Meanwhile, our method generates dense depth maps using LiDAR, a process that is time-consuming. Although we have compared several depth completion methods for reference, challenges persist: propagation-based approaches, despite their high accuracy, suffer from excessively long inference time; conversely, deep learning-based methods, while enabling real-time rendering, face issues of prolonged neural network training and poor generalization capability.

Future work should prioritize incorporating accelerated training, LoD rendering techniques, and efficient and accurate depth completion methods.

## 5. Conclusion

This study introduces ARSGaussian, a novel view synthesis method for aerial remote sensing that leverages LiDAR point cloud as a constraint. Our approach not only achieves realistic novel view synthesis in large-scale aerial remote sensing scenes but also recovers high-precision geometric structures which is essential for remote sensing. Firstly, our method incorporates LiDAR point cloud as constraints for Gaussian point densification, guiding the growth and splitting of Gaussians along the geometric benchmarks provided by LiDAR point cloud. This addresses the issue of floaters existing in aerial remote sensing scenes, thereby enhancing the visual quality of synthesized views. Secondly, the method derives coordinate transformations including distortion parameters for camera models to achieve pixel-level alignment between LiDAR point cloud and 2D images, enabling the fusion of heterogeneous data and high-precision geo-alignment. Finally, our approach integrates geometric consistency loss into the regularization process, encouraging Gaussians to align closer to the real depth and plane representations, which significantly improves the geometric estimation accuracy of the model. Experiments demonstrate that our method excels in various visual quality metrics and substantially enhances precision in geometric estimation.

## Acknowledgments

This work was supported by The National Key Research and Development Program of China（No.2021YFB3900500）, The Science and Disruptive Technology Program, AIRCAS（No.2024-AIRCAS-SDTP-11） and The Fundamental Enhancement Program in Technical Fields (No.: 2024-JCJQ-JJ-0333).

The authors would like to thank all the researchers who kindly shared the codes used in this paper. We would also like to thank all the research organizations that provided the datasets in this paper.